\documentclass{article} 
\usepackage{iclr2025_conference,times}
\usepackage{hyperref}
\usepackage{url}

\usepackage{multirow}
\usepackage{hhline}
\usepackage{tabularx}
\usepackage{amsthm}
\newtheorem{theorem}{Theorem}

\usepackage{algorithm, setspace} 
\usepackage{algpseudocode} 
\usepackage{adjustbox}
\usepackage{subfig}
\usepackage{booktabs}
\usepackage{enumitem}
\usepackage{makecell}

\usepackage{thmtools}
\usepackage{thm-restate}
\usepackage{hyperref}
\usepackage{amsmath}
\usepackage{cleveref}

\usepackage{tabularx} 
\usepackage{siunitx,booktabs} 
\usepackage{amssymb}
\iclrfinalcopy

\usepackage{wrapfig}
\usepackage{tcolorbox}

\usepackage{amsmath,amsfonts,bm}

\def\eqref#1{equation~\ref{#1}}

\def\1{\bm{1}}

\DeclareMathAlphabet{\mathsfit}{\encodingdefault}{\sfdefault}{m}{sl}
\SetMathAlphabet{\mathsfit}{bold}{\encodingdefault}{\sfdefault}{bx}{n}

\usepackage{hyperref}
\usepackage{url}

\usepackage{colortbl}

\title{PaCA: Partial Connection Adaptation \\ for Efficient Fine-Tuning}

\author{Sunghyeon Woo, Sol Namkung, Sunwoo Lee, Inho Jeong, Beomseok Kim, Dongsuk Jeon \\
Seoul National University\\
\texttt{\{wsh0917, djeon1\}@snu.ac.kr}
}

\begin{document}

\maketitle

\begin{abstract}
Prior parameter-efficient fine-tuning (PEFT) algorithms reduce memory usage and computational costs of fine-tuning large neural network models by training only a few additional adapter parameters, rather than the entire model. However, the reduction in computational costs due to PEFT does not necessarily translate to a reduction in training time; although the computational costs of the adapter layers are much smaller than the pretrained layers, it is well known that those two types of layers are processed sequentially on GPUs, resulting in significant latency overhead. LoRA and its variants avoid this latency overhead by merging the low-rank adapter matrices with the pretrained weights during inference. However, those layers cannot be merged during training since the pretrained weights must remain frozen while the low-rank adapter matrices are updated continuously over the course of training. Furthermore, LoRA and its variants do not reduce activation memory, as the first low-rank adapter matrix still requires the input activations to the pretrained weights to compute weight gradients. To mitigate this issue, we propose \textbf{Pa}rtial \textbf{C}onnection \textbf{A}daptation (\textbf{PaCA}), which fine-tunes randomly selected partial connections within the pretrained weights instead of introducing adapter layers in the model. PaCA not only enhances training speed by eliminating the time overhead due to the sequential processing of the adapter and pretrained layers but also reduces activation memory since only partial activations, rather than full activations, need to be stored for gradient computation. Compared to LoRA, PaCA reduces training time by 22\% and total memory usage by 16\%, while maintaining comparable accuracy across various fine-tuning scenarios, such as fine-tuning on the MMLU dataset and instruction tuning on the Oasst1 dataset. PaCA can also be combined with quantization, enabling the fine-tuning of large models such as LLaMA3.1-70B. In addition, PaCA enables training with 23\% longer sequence and improves throughput by 16\% on both NVIDIA A100 GPU and INTEL Gaudi2 HPU compared to LoRA. The code is available at {\href{https://github.com/WooSunghyeon/paca}{https://github.com/WooSunghyeon/paca}}.

\end{abstract}

\section{Introduction} \label{sec: introduction}

Following the scaling laws \citep{Kaplan-scaling-laws1, Hoffmann-scaling-laws2}, the size of language models based on the transformer architecture \citep{Vaswani-transformer} has grown significantly in recent years. Large Language Models (LLMs) such as GPT4 \citep{GPT-4-OPENAI} and LLaMA 3 \citep{LLaMA3-Dubey} have achieved remarkable abilities across a wide range of general tasks. Furthermore, the capabilities of LLMs can be refined for specific purposes, either by creating models specialized for specific tasks through fine-tuning \citep{singhal-parm2} or by developing chatbots that better understand user queries through instruction tuning \citep{Wei-instruction1-flan, taori-insturction2-alpaca}. However, fine-tuning LLMs consumes significant computational power and memory, making it impossible to perform without a large number of expensive GPUs.

Parameter-efficient fine-tuning (PEFT) \citep{prefix-Lisa, series-adapter-Houlsby, parallel-adapter-He} is a set of methods to relieve the high costs of fine-tuning large models. Prior PEFT schemes introduce new adapter layers with significantly fewer parameters to a pretrained model and only train these newly introduced adapter layers, substantially reducing the memory needed to store gradients and optimizer states. Furthermore, PEFT can reduce the computational overhead of fine-tuning, as it needs to calculate the parameter gradients only for the adapter weights, rather than for all model parameters.

However, we observed that the reduction in computational cost due to PEFT does not translate into a significant decrease in actual training time. This issue arises from the fact that the adapter layers are typically processed sequentially with the pretrained layers since GPUs are generally optimized for processing one kernel at a time. This sequential processing limits the full utilization of hardware resources and incurs significant latency overhead, even though the number of FLOPs of the adapter layers is significantly smaller than that of the pretrained layers. While some software tools such as CUDA streams could be used to process the adapter layers in parallel by executing multiple kernels simultaneously, it suffers from the overhead of managing and synchronizing the streams \citep{wang-multikernel1, dai-multikernel2, han-multikernel3}.

LoRA \citep{LoRA-Hu} and its variants \citep{Vera-Kopiczko, DoRA-Yang, MosLoRA-Wu} avoid this latency overhead by merging the low-rank adapter matrices and the pretrained weights to eliminate the need for sequential processing during inference. However, this approach cannot be applied to fine-tuning since the low-rank adapter matrices need to be trained separately from the frozen pretrained weights, making the overhead from sequential processing unavoidable. Furthermore, LoRA and its variants do not reduce activation memory compared to Full-FT, since the input activations of the pretrained weights still need to be stored in memory to calculate the gradients for the first low-rank adapter matrix.

In this paper, we propose \textbf{PaCA} (\textbf{Pa}rtial \textbf{C}onnection \textbf{A}daptation), which fine-tunes randomly selected partial connections in the pretrained weights without relying on adapter layers, as depicted in Fig. \ref{fig: PaCA}. Unlike prior PEFT schemes, PaCA successfully reduces training time since the forward and backward operations for the pretrained weights also include those for the partial connections, eliminating the need for additional sequential processing. Furthermore, since calculating the gradients for the partial weights only requires the corresponding activations, PaCA significantly reduces activation memory usage as well. We first theoretically show that PaCA can effectively converge the loss in general neural networks. In experiments with various scenarios, PaCA demonstrates substantial reductions in both training time and memory compared to prior PEFT schemes while maintaining comparable accuracy on NVIDIA A100 GPU \citep{a100-choquette} and Intel Gaudi2 HPU \citep{gaudi2-interl}. In summary, our contributions are as follows:

 \begin{figure}[!t]
  \centering
  \includegraphics[width=0.75\linewidth]{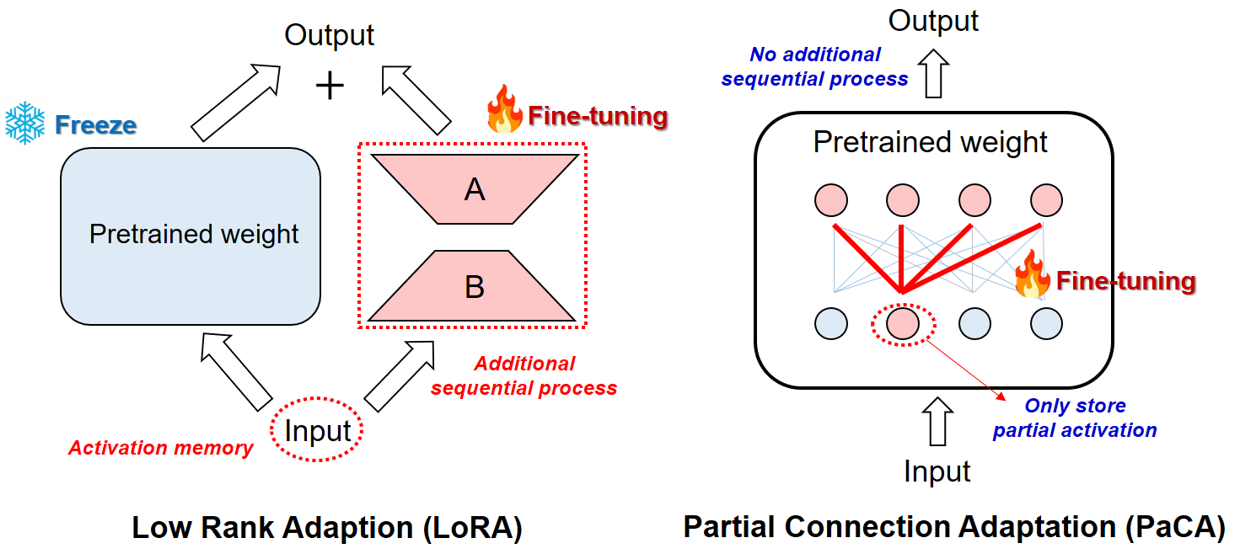}
  \caption{Overview of Partial Connections Adaptation (PaCA) algorithm.} 
  \label{fig: PaCA}
\end{figure}

\begin{itemize}
\item We propose PaCA, a memory-efficient PEFT algorithm that fine-tunes randomly selected partial connections within pretrianed weights without using additional adapter layers.
\item We theoretically prove that PaCA can converge the loss in general neural networks.
\item We experimentally show that PaCA effectively reduces memory consumption and improves training speed compared to prior PEFT algorithms across various fine-tuning scenarios on different types of GPUs.
\end{itemize}

\section{Background \& Motivation} \label{sec: background & motivation}

In general, training deep neural networks involves backpropagation \citep{rumelhart1-backprop}, which facilitates the adaptation of the model in the direction that minimizes the loss function. The equations below show the backpropagation algorithm for a linear layer:

\begin{align}
 \text{Forward:} \quad &\textbf{X}_{out} = \textbf{W} \textbf{X}_{in} \label{eq: forward propagation} \\
 \text{Backward:} \quad &\nabla \textbf{X}_{in} =  \textbf{W}^{T} \nabla\textbf{X}_{out}  \label{eq: backward propagation} \\
  &\nabla \textbf{W} = \nabla \textbf{X}_{out} \, \textbf{X}_{in}^{T} \label{eq: weight gradient}
 \end{align}

where $\textbf{W}\in\mathbb{R}^{d_{out}\times d_{in}} $, $\textbf{X}_{in} \in \mathbb{R}^ {d_{in}} $, and $\textbf{X}_{out} \in \mathbb{R}^ {d_{out}} $ denote the weights, input activations, and output activations, respectively, with $d_{in}$ and $d_{out}$ denoting the input and output dimensions of the layer. $\nabla \textbf{W}$ and $\nabla \textbf{X}_{in}$ represent the weight gradients and input gradients. The forward propagation computes the output activations following Eq. \ref{eq: forward propagation}, while the backward propagation computes the input gradients (Eq. \ref{eq: backward propagation}) and the weight gradients (Eq. \ref{eq: weight gradient}).  

Full-FT trains all layers using backpropagation, performing the operations described in Eqs. \ref{eq: forward propagation}-\ref{eq: weight gradient} for each layer. Consequently, Full-FT incurs significant memory overhead due to storing the gradients and optimizer states for all parameters. To lower this overhead, various PEFT schemes have been introduced. For instance, the training scheme of LoRA \citep{LoRA-Hu}, a representative PEFT algorithm, is represented as the equations below:

\begin{align}
 \text{Forward:} \quad &\textbf{X}_{out} = \textbf{W} \textbf{X}_{in} + \textcolor{blue}{\textbf{B} (\textbf{A} \textbf{X}_{in})} \label{eq: lora forward propagation} \\
 \text{Backward:} \quad &\nabla \textbf{X}_{in} =  \textbf{W}^{T} \nabla\textbf{X}_{out} + \textcolor{blue}{\textbf{A}^{T} (\textbf{B}^{T} \nabla\textbf{X}_{out})} \label{eq: lora backward propagation} \\
  &\textcolor{blue}{\nabla \textbf{B} = \nabla \textbf{X}_{out} \, \textbf{X}_{mid}^{T}} \, , \; \textcolor{blue}{\nabla \textbf{A} = \nabla \textbf{X}_{mid} \, \textbf{X}_{in}^{T}}  \label{eq: lora weight gradient}
 \end{align}

where $\textbf{B} \in \mathbb{R}^{d_{out} \times r}$ and $\textbf{A} \in \mathbb{R}^{r \times d_{in}}$ represent the low-rank adapter matrices in LoRA, with $r$ denoting the rank of the adapter. $\textbf{X}_{mid} \in \mathbb{R}^{r}$ represents the output activations after propagating through the LoRA $\textbf{A}$ layer (i.e., $\textbf{X}_{mid} = \textbf{A} \textbf{X}_{in}$). In Eqs. \ref{eq: lora forward propagation}-\ref{eq: lora weight gradient}, we have highlighted the computations involving adapter weights in blue. Compared to Full-FT, prior PEFT schemes introduce two key changes: 1) computations for the adapters are added in forward and backward propagations (Eqs. \ref{eq: lora forward propagation}-\ref{eq: lora backward propagation}), and 2) only the adapters are trained, excluding the pretrained weights (Eq. \ref{eq: lora weight gradient}). Since the computational cost of the adapters in PEFT is typically negligible compared to that of the pretrained layers \citep{prefix-Lisa, series-adapter-Houlsby, parallel-adapter-He, LoRA-Hu}, PEFT can reduce the overall computational cost of training by eliminating the need to compute parameter gradients for the pretrained weights.

\begin{figure}[!t]
\centering
\subfloat[Operations per iteration. \label{fig: FLOPs per iteration.}]{\includegraphics[width=.33\textwidth]{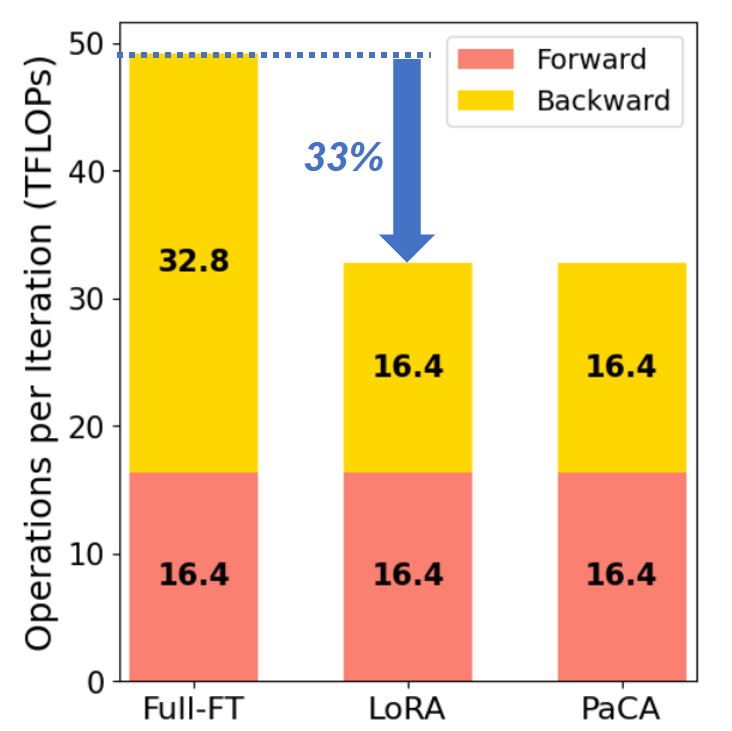}} \qquad \quad
\subfloat[Training time per iteration. \label{fig: Time per iteration.}]{\includegraphics[width=.33\textwidth]{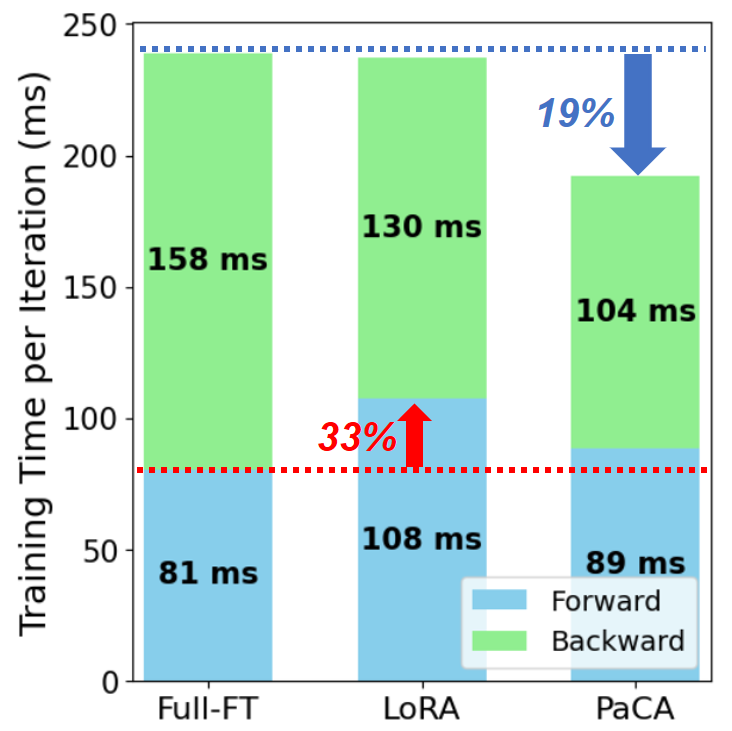}}
\caption{The number of operations (TFLOPs) and training time (ms) per iteration when training LLaMA3-8B with full-fine tuning (Full-FT) and LoRA.}
\label{fig: introduction}
\end{figure}

For more detailed analysis, we calculate FLOPs and measure training time when fine-tuning the LLaMA3-8B model using Full-FT and LoRA. Experimental results show that the operation count of LoRA is approximately 33\% lower than Full-FT (Fig. \ref{fig: FLOPs per iteration.}). However, the saving in actual training time is only 0.6\%, as displayed in Fig. \ref{fig: Time per iteration.}, which is far below the expected 33\% decrease. To investigate this discrepancy, we analyzed the computational cost for both forward and backward propagation, as well as the actual training time. 

One interesting finding is that the time required for forward propagation in LoRA increased by 33\% compared to Full-FT, despite requiring a similar number of operations, as shown in Fig. \ref{fig: Time per iteration.}. This latency overhead is due to the inefficient sequential processing of the pretrained and adapter layers, as reported by \cite{LoRA-Hu}. More specifically, the operations associated with the adapter layers are conventionally executed in a sequential manner, rather than in parallel with the pretrained layers, as GPUs are typically designed to execute a single kernel at a time. Although parallel execution of the adapter layers may be feasible using CUDA streams, which allow multiple kernels to run concurrently, these methods introduce additional overhead of resource allocation and synchronization between streams \citep{wang-multikernel1, dai-multikernel2, han-multikernel3}.

This sequential processing of the adapter and pretrained layers negatively impacts hardware utilization and incurs latency overhead, despite the fact that the computational cost of the adapter layers accounts for only approximately 1\% of that of the pretrained layers. This latency overhead could be mitigated by merging the low-rank adapter matrices into the pretrained weights during inference \citep{LoRA-Hu}. However, during fine-tuning, where the pretrained weights must remain frozen and only the adapter weights are updated separately, such merging is not possible and the latency overhead from sequential processing remains. 

Furthermore, LoRA and its variants are unable to reduce the activation memory. In Full-FT, all input activations ($\textbf{X}_{in}$) must be stored in memory during forward propagation in order to calculate the gradients of the pretrained weights ($\nabla \textbf{W}$) in backward propagation, as shown in Eq. \ref{eq: weight gradient}. Although LoRA does not require the computation of gradients for the pretrained weights, the input activations ($\textbf{X}_{in}$) must still be stored in memory to calculate the gradients for the LoRA \textbf{A} layer ($\nabla \textbf{A}$), as indicated in Eq. \ref{eq: lora weight gradient}. Additionally, the output activations of the LoRA \textbf{A} layer ($\textbf{X}_{mid}$) must be stored in memory to calculate the gradients for the LoRA \textbf{B} layer ($\nabla \textbf{B}$) following Eq. \ref{eq: lora weight gradient}. This issue with activation memory becomes more critical when training on long sequence data or increasing batch size to improve training throughput \citep{DropIT-Chen, Megatron-Korthikanti, ALAM-Woo}.

\section{Methodology} \label{sec: methdology}

\subsection{PaCA: Partial Connection Adaptation} \label{sec: paca: partial connection adaptation}

Motivated by the observation that the newly introduced adapter layers lead to training inefficiencies, we propose \textbf{Pa}rtial \textbf{C}onnection \textbf{A}daptation (\textbf{PaCA}). PaCA fine-tunes randomly selected partial connections within the pretrained weights rather than introducing new adapter layers, as depicted in Fig. \ref{fig: PaCA}. More specifically, PaCA employs the training algorithm below:

\begin{align}
 \text{Forward:} \quad &\textbf{X}_{out} = \textbf{W} \textbf{X}_{in} \label{eq: paca forward propagation} \\
 \text{Backward:} \quad &\nabla \textbf{X}_{in} =  \textbf{W}^{T}  \nabla\textbf{X}_{out}  \label{eq: paca backward propagation} \\
  & \textcolor{black}{\nabla \textbf{P} = \nabla \textbf{X}_{out} \, {}^{p}{\textbf{X}}_{in}^{T}} \label{eq: paca weight gradient}
 \end{align}

where $\textbf{P} \in \mathbb{R}^{d_{out} \times r}$ and ${}^{p}{\textbf{X}}_{in} \in \mathbb{R}^{r}$ denote the partial connections randomly selected from the pretrained weights (i.e., $\textbf{P} \subset \textbf{W}$) and the corresponding partial activations selected from the input activations (i.e., ${}^{p}{\textbf{X}}_{in} \subset \textbf{X}_{in}$), respectively. $r$ represents the number of the randomly selected columns within the pretrained weights, which we refer to \textit{rank} when PaCA is applied. The operations involving partial connections are highlighted in red. 

PaCA randomly selects the partial connections to fine-tune from the pretrained weights before training and then fine-tunes only the selected connections. Since these partial connections are part of the pretrained weights, no additional computations are required in forward and backward computations (Eqs. \ref{eq: paca forward propagation}-\ref{eq: paca backward propagation}), completely avoiding inefficient sequential processing due to the adapter layers in LoRA. In addition, while LoRA requires both the input activations ($\textbf{X}_{in}$) and the output activations of the LoRA \textbf{A} layer ($\textbf{X}_{mid}$) to calculate gradients for the low-rank adapter matrices (Eq. \ref{eq: lora weight gradient}), PaCA only needs to store the partial activations (${}^{p}{\textbf{X}}_{in}$) to calculate the gradients of the partial connections ($\nabla \textbf{P}$), significantly reducing the amount of activation to be temporarily stored in memory.

We calculated the FLOPs and measured the training time required for fine-tuning the LLaMA3-8B model using PaCA to demonstrate its effectiveness (see Table \ref{tb: hyperparameters, FLOPs vs actual time} in Appendix \ref{appendix: experimental details} for experiment details), and the results are summarized in Fig. \ref{fig: introduction}. Experimental results indicate that PaCA provides a 19\% reduction in total training time compared to LoRA, by reducing forward propagation time by 18\% and backward propagation time by 20\%, achieved through avoiding additional sequential processing. One interesting observation is that while the FLOPs required for forward and backward propagation in PaCA are nearly identical, the actual runtime for backward propagation is 17\% longer than forward propagation. We hypothesize that, even though the computation of weight gradients for partial connections (Eq. \ref{eq: paca weight gradient}) is significantly smaller than that for the pretrained weights, it occurs sequentially with the input gradient computation (Eq. \ref{eq: paca backward propagation}) during backward propagation. This sequential processing introduces additional latency compared to forward propagation, which only involves the computation of output activations (Eq. \ref{eq: forward propagation}). It should be noted that this latency overhead is not a specific overhead introduced by PaCA, but rather an inherent issue in all backpropagation-based training algorithms including Full-FT and prior PEFT algorithms, which must compute both input gradients and weight gradients. 

\textcolor{black}{Intuitively, training only a subset of connections can be interpreted as learning within a subspace composed of the selected connections. Prior studies revealed that overparameterized models can be efficiently trained even when weights are projected onto a small subspace \citep{Li-subspace1, Aghajanyan-subspace2}. Similarly, LoRA \citep{LoRA-Hu} was suggested based on the assumption that weight updates can be projected onto a small low-rank subspace. Inspired by these observations, we hypothesized that weight updates could also be projected onto a small subspace composed of a subset of weight columns. In other words, we assumed that the critical factor is learning within a small subspace, not the method of selecting the subspace itself. Here we prove that training only a subset of connections is sufficient to ensure the convergence of loss in neural networks, as demonstrated in Section \ref{sec: convergence analysis of paca}.}

\subsection{Convergence Analysis of PaCA} \label{sec: convergence analysis of paca}

In Section \ref{sec: paca: partial connection adaptation}, we proposed PaCA and demonstrated its effectiveness. Now we theoretically prove that PaCA converges for general neural networks. We first define the input at the $k$-th iteration as $\textbf{X}^{k}$ and the full set of weights as $\textbf{W}^{k} = [\textbf{W}_{1}^{k}, \textbf{W}_{2}^{k}, \dots, \textbf{W}_{n}^{k}]$, where $n$ denotes the number of layers. The loss of the model is defined as $f(\textbf{X}^{k}, \textbf{W}^{k})$. The weight of the $l$-th layer $\textbf{W}_{l}^{k}$ can be represented as a collection of column vectors (i.e., $\textbf{W}_{l}^{k} = [{}_{1}\textbf{w}_{l}^{k}, {}_{2}\textbf{w}_{l}^{k}, \dots, {}_{d_{l}}\textbf{w}_{l}^{k}]$). In PaCA, we only fine-tune randomly selected columns $\textbf{P}_{l}^{k} = [{}_{i_{1}}\textbf{w}_{l}^{k}, {}_{i_{2}}\textbf{w}_{l}^{k}, \dots, {}_{i_{r}}\textbf{W}_{l}^{k}]$ where $i_{1}, \dots, i_{r}$ denote the selected column indices for PaCA. The weights are then updated as follows:

\begin{align}
 \text{Full-FT:}\,\,\,\,\,\, &\textbf{W}_{l}^{k+1} = \textbf{W}_{l}^{k} - \eta \nabla \textbf{W}_{l}^{k} = \textbf{W}_{l}^{k} - \eta [\nabla{}_{1}\textbf{w}_{l}^{k}, \nabla{}_{2}\textbf{w}_{l}^{k}, \dots, \nabla{}_{d_{l}}\textbf{w}_{l}^{k}] \label{eq: fft weight update} \\
 \text{PaCA:}\,\,\,\,\,\, &\textbf{W}_{l}^{k+1} = \textbf{W}_{l}^{k} - \eta \Delta \textbf{W}_{l}^{k} = \textbf{W}_{l}^{k} - \eta [\textbf{0}, \nabla{}_{i_{1}}\textbf{w}_{l}^{k}, \dots,\nabla{}_{i_{r}}\textbf{w}_{l}^{k},\dots  \textbf{0}] \label{eq: paca weight update}
 \end{align}

 where $\eta$ denotes learning rate and $\Delta \textbf{W}_{l}^{k}$ denotes weight updates. In this scenario, we define the full set of partial connections within the model as $\textbf{P}^{k}=[\textbf{P}_{1}^{k}, \textbf{P}_{2}^{k}, \dots, \textbf{P}_{n}^{k}]$. Then, PaCA satisfies the following theorem:

\begin{theorem}\label{theorem1: paca convergence} 
If the gradient of the loss function $f(\textbf{W}, \textbf{X})$ is Lipschitz continuous and the only partial connections are updated, then 
\begin{equation*}
    f(\textbf{W}^{k+1}, \textbf{X}^{k+1}) \leq f(\textbf{W}^{k}, \textbf{X}^{k}) -\eta(1-\frac{\eta L}{2}) ||\nabla \textbf{P}^{k}||^{2}
\end{equation*}
\end{theorem}

We prove Theorem \ref{theorem1: paca convergence} by applying Eq. \ref{eq: paca weight update} to the quadratic upper bound using Lipschitz continuity condition (i.e., $f(\textbf{W}^{k+1}, \textbf{X}^{k+1}) \leq f(\textbf{W}^{k}, \textbf{X}^{k}) + \nabla_{\textbf{W}^{k}}f(\textbf{W}^{k}, \textbf{X}^{k})(\textbf{W}^{k+1}-\textbf{W}^{k})^{T}+L/2||\textbf{W}^{k+1}-\textbf{W}^{k}||^{2}$) where $L$ denotes the Lipschitz constant. The detailed proof can be found in Appendix \ref{appendix: proof for the convergence of PaCA}. Theorem \ref{theorem1: paca convergence} implies that as long as the learning rate $\eta$ is chosen to satisfy the condition $0<\eta<2/L$, the loss function $f(\textbf{W}, \textbf{X})$ will decrease after each iteration, ensuring convergence of the neural network. 



\section{Experiments} \label{sec: experiments}

To verify the effectiveness of PaCA, here we evaluate its performance in various fine-tuning scenarios. Section \ref{sec: task specific fine-tuning} first compares the accuracy and performance of PaCA with other PEFT algorithms, such as LoRA \citep{LoRA-Hu}, DoRA \citep{DoRA-Yang}, and MosLoRA \citep{MosLoRA-Wu}, when fine-tuning the LLaMA2-7B/13B \citep{llama2-tourvron} and LLaMA3-8B \citep{LLaMA3-Dubey} models on the MMLU dataset \citep{mmlu-hendryckstest2021}. In Section \ref{sec: instruction tuning}, we observe the instruction-following ability on the MT-Bench dataset \citep{mt-bench-zheng} after fine-tuning the LLaMA3-8B model with PaCA and the LoRA family on the Oasst1 dataset \citep{oasst1-kopf}. In Section \ref{sec: qpac: enhancements to qlora}, we compare the performance and score of our quantized PaCA (QPaCA) with QLoRA \citep{qlora-dettmers} on the MT-Bench \citep{mt-bench-zheng} dataset while fine-tuning the LLaMA3.1-70B \citep{LLaMA3-Dubey} model on the Oasst1 dataset. Section \ref{sec: training performance of paca} analyzes the ability of PaCA and the LoRA family to handle long sequence data and the training throughput when increasing the batch size, using both a single NVIDIA A100 \citep{a100-choquette} and Intel Gaudi2 HPU \citep{gaudi2-interl}. \textcolor{black}{In addition, we tested PaCA on different model architectures such as the vision transformer (ViT \citep{vit-dosovitskiy}) and convolutional neural network (EfficientNet-V2 \citep{efficientnetv2-tan}) for demonstrating generalizability of PaCA in Appendix \ref{appendix: vision} } 

\subsection{Fine-Tuning for Specific Tasks} \label{sec: task specific fine-tuning}
\begin{table}[!b]
\caption{Comparisons of memory usage (Mem), training time (Time), and 5-shot accuracy on MMLU dataset when fine-tuning LLaMA2-7B/13B and LLaMA3-8B models using various PEFT algorithms. Param indicates the number of trainable parameters.}
\centering
\label{tb: task specific fine-tuning}
\renewcommand{\arraystretch}{1.1} 
\resizebox{\textwidth}{!}{%
\begin{tabular}{c|c|c c|c c|cccc|c}
\toprule
\multirow{2}{*}[-0.6ex]{\textbf{Model}} & \multirow{2}{*}[-0.6ex]{\textbf{Method}} & \multirow{2}{*}[-0.6ex]{\textbf{Rank}} & \multirow{2}{*}[-0.6ex]{\textbf{Param}} & \multirow{2}{*}[-0.6ex]{\textbf{Mem}} & \multirow{2}{*}[-0.6ex]{\textbf{Time}} & \multicolumn{5}{c}{\textbf{Accuracy (\%)}} \\ 
\cmidrule(lr){7-11}
 &  &  &  &  &  & \textbf{Hums.} & \textbf{STEM} & \textbf{Social.} & \textbf{Other} & \textbf{Avg.} \\ 
\midrule
\multirow{6}{*}{\textbf{LLaMA2-7B}} & No tuning & - & - & - & - & 44.0 & 37.0 & 51.5 & 53.1 & 45.9 \\
 & LoRA & 8 & 20M & 23G & 4.1h & 48.5 & 41.2 & 57.3 & 56.5 & 50.6 \\ 
 & DoRA & 8 & 21M & 29G & 8.7h & 48.7 & 42.3 & 58.3 & 57.6 & \textbf{51.3} \\ 
 & MosLoRA & 8 & 20M & 23G & 4.3h & 46.6 & 42.2 & 60.8 & 57.4 & 51.1 \\ 
 & \cellcolor{blue!10}  & \cellcolor{blue!10} 8 & \cellcolor{blue!10} 11M & \cellcolor{blue!10} \textbf{20G} & \cellcolor{blue!10} \textbf{3.2h} & \cellcolor{blue!10} 46.8 & \cellcolor{blue!10} 41.1 & \cellcolor{blue!10} 58.4 & \cellcolor{blue!10} 57.3 & \cellcolor{blue!10} 50.4 \\ 
 & \cellcolor{blue!10} \multirow{-2}{*}{PaCA (Ours)}  & \cellcolor{blue!10} 16 & \cellcolor{blue!10} 22M & \cellcolor{blue!10} \textbf{20G} & \cellcolor{blue!10} \textbf{3.2h} & \cellcolor{blue!10} 48.7 & \cellcolor{blue!10} 41.7 & \cellcolor{blue!10} 58.7 & \cellcolor{blue!10} 57.6 & \cellcolor{blue!10} 51.2 \\ 
\midrule
\multirow{6}{*}{\textbf{LLaMA2-13B}} & No tuning & - & - & - & - & 53.1 & 44.2 & 62.8 & 60.8 & 54.9 \\ 
 & LoRA & 8 & 31M & 40G & 6.3h & 53.9 & 46.2 & 66.8 & 62.9 & 57.0 \\ 
 & DoRA & 8 & 33M & 49G & 14.7h & 55.6 & 46.8 & 66.7 & 64.8 & \textbf{58.1} \\ 
 & MosLoRA & 8 & 31M & 40G & 6.5h & 56.5 & 47.3 & 66.1 & 62.8 & 57.9 \\ 
 & \cellcolor{blue!10} & \cellcolor{blue!10} 8 & \cellcolor{blue!10} 17M & \cellcolor{blue!10} \textbf{35G} & \cellcolor{blue!10} \textbf{5.2h} & \cellcolor{blue!10} 52.7 & \cellcolor{blue!10} 46.2 & \cellcolor{blue!10} 67.1 & \cellcolor{blue!10} 63.4 & \cellcolor{blue!10} 56.8 \\ 
 & \cellcolor{blue!10} \multirow{-2}{*}{PaCA (Ours)}  & \cellcolor{blue!10} 16 & \cellcolor{blue!10} 34M & \cellcolor{blue!10} \textbf{35G} & \cellcolor{blue!10} \textbf{5.2h} & \cellcolor{blue!10} 56.0 & \cellcolor{blue!10} 46.7 & \cellcolor{blue!10} 66.3 & \cellcolor{blue!10} 64.0 & \cellcolor{blue!10} 58.0 \\ 
\midrule
\multirow{6}{*}{\textbf{LLaMA3-8B}} & No tuning & - & - & - & - & 59.3 & 55.3 & 75.7 & 72.7 & 64.9 \\ 
 & LoRA & 8 & 21M & 27G & 4.4h & 59.4 & 56.3 & 75.4 & 71.9 & 65.0 \\ 
 & DoRA & 8 & 22M & 33G & 9.4h & 59.4 & 56.3 & 75.7 & 72.2 & 65.2 \\ 
 & MosLoRA & 8 & 21M & 27G & 4.6h & 59.8 & 55.9 & 75.7 & 72.0 & 65.1 \\ 
 &  \cellcolor{blue!10}  & \cellcolor{blue!10} 8 & \cellcolor{blue!10} 11M & \cellcolor{blue!10} \textbf{23G} & \cellcolor{blue!10} \textbf{3.5h} & \cellcolor{blue!10} 59.7 & \cellcolor{blue!10} 55.7 & \cellcolor{blue!10} 76.0 & \cellcolor{blue!10} 72.3 & \cellcolor{blue!10} 65.2 \\ 
 & \multirow{-2}{*}{\cellcolor{blue!10} PaCA (Ours)}  & \cellcolor{blue!10} 16 & \cellcolor{blue!10} 22M & \cellcolor{blue!10} \textbf{23G} & \cellcolor{blue!10} \textbf{3.5h} & \cellcolor{blue!10} 60.2 & \cellcolor{blue!10} 55.9 & \cellcolor{blue!10} 75.8 & \cellcolor{blue!10} 72.6 & \cellcolor{blue!10} \textbf{65.4} \\ 
\bottomrule
\end{tabular}%
}
\end{table}

We first compared PaCA against LoRA, DoRA, and MosLoRA using the MMLU dataset, which consists of 57 tasks designed to assess the ability of a model to understand and reason across a wide range of academic subjects \citep{mmlu-hendryckstest2021}. The evaluation was conducted on the LLaMA2-7B/13B and LLaMA3-8B models, with the rank of the prior PEFT methods set to 8. We employ PaCA with a rank of 8 and 16, each representing the case where the rank is equal to that of prior PEFT methods and where the number of trainable parameters is identical. Aside from adjusting the learning rate for each PEFT model, all other experimental settings remained identical, as detailed in Table \ref{tb: hyperparameters, mmlu} in Appendix \ref{appendix: experimental details}. All experiments were conducted on a single NVIDIA A100 GPU.

The experimental results in Table \ref{tb: task specific fine-tuning} demonstrate that PaCA significantly reduces both memory usage and training time across all models, while maintaining accuracy comparable to the other PEFT algorithms. For the LLaMA2-7B model, PaCA achieves accuracy similar to LoRA when the rank is set to 8, despite using only half the number of trainable parameters, reducing memory usage by 13\% and training time by 26\% simultaneously. In this configuration, the accuracy of PaCA drops by up to 0.9\% compared to LoRA variants such as DoRA and MosLoRA. However, when the rank of PaCA is increased to 16, matching the number of trainable parameters with DoRA and MosLoRA, PaCA achieves almost identical accuracy to DoRA and MosLoRA while still offering considerable reductions in memory usage and training time. Specifically, PaCA reduces memory usage by 31\% and training time by 63\% compared to DoRA, while offering a 13\% reduction in memory usage and a 26\% reduction in training time compared to MosLoRA.

A similar trend is observed in both LLaMA2-13B and LLaMA3-8B, where PaCA continues to show substantial reductions in memory usage and training time. On LLaMA2-13B, PaCA achieves comparable accuracy to the LoRA variants while reducing memory usage by 13\%, 29\%, and 13\%, and training time by 17\%, 64\%, and 20\%, compared to LoRA, DoRA, and MosLoRA, respectively. In LLaMA3-8B, PaCA consumes the least memory and training time among LoRA and its variants, while achieving the highest accuracy. In summary, PaCA successfully improves training speed by eliminating unnecessary sequential processes and reduces memory usage by storing only partial activations, while maintaining comparable accuracy in fine-tuning scenarios on specific tasks.

\subsection{Instruction Tuning} \label{sec: instruction tuning}
We next evaluate PaCA on the MT-Bench dataset, which consists of 80 queries designed to measure the instruction-following capabilities of a model across multiple tasks, providing a detailed assessment of its performance in real-world scenarios \citep{mt-bench-zheng}. Specifically, we fine-tuned LLaMA3-8B using a single NVIDIA A100 GPU on the Oasst1 dataset, which is an instruction-following dataset, and then evaluated the score on the MT-Bench dataset using GPT4o-mini as the judge. The detailed setup can be found in Table \ref{tb: hyperparameters, oasst1} in Appendix \ref{appendix: experimental details}. 

\begin{table}[!htb]
\caption{Comparisons of memory usage (Mem), training time (Time), and score on MT-Bench dataset when fine-tuning LLaMA3-8B on Oasst1 dataset using various PEFT algorithms.}
\label{tb: insturction-tuning}
\centering
\renewcommand{\arraystretch}{1.1}
\resizebox{\textwidth}{!}{%
\begin{tabular}{c|c|cc|cccccccc|c}
\toprule
\textbf{Method} & \textbf{Rank} & \textbf{Mem} & \textbf{Time} & \textbf{Human.} & \textbf{STEM} & \textbf{Role.} & \textbf{Extract.} & \textbf{Writing} & \textbf{Reason.} & \textbf{Coding} & \textbf{Math} & \textbf{Avg.} \\ 
\midrule
No tuning & - & - & - & 6.25 & 5.70 & 5.45 & 4.85 & 5.20 & 4.40 & 3.20 & 1.95 & 4.62 \\
LoRA &  64 & 56G & 26m & 7.00 & 6.40 & 5.70 & 5.80 & 5.30 & 4.55 & 3.25 & 2.95 & 5.12 \\ 
DoRA &  64 & 65G & 50m & 6.95 & 6.00 & 5.90 & 5.80 & 6.20 & 4.50 & 3.50 & 3.40 & \textbf{5.28} \\ 
MosLoRA & 64 & 56G & 27m & 6.90 & 6.50 & 5.80 & 5.70 & 5.55 & 4.90 & 3.10 & 2.75 & 5.15 \\ 
\rowcolor{blue!10}  & 64 & \textbf{47G} & \textbf{21m} & 6.50 & 6.30 & 5.90 & 5.95 & 5.65 & 4.80 & 3.70 & 3.05 & 5.23 \\ 
\rowcolor{blue!10} \multirow{-2}{*}{ PaCA (Ours)}  & 128 & 51G & \textbf{21m} & 6.80 & 6.15 & 6.05 & 5.95 & 5.85 & 4.65 & 3.45 & 3.15 & 5.26 \\ 
\bottomrule
\end{tabular}%
}
\end{table}

Table \ref{tb: insturction-tuning} confirms that PaCA significantly reduces memory usage and training time compared to other PEFT methods while maintaining comparable scores, consistent with the results observed when fine-tuning it on the MMLU dataset. Specifically, our PaCA outperforms LoRA and MosLoRA with 16\% less memory usage and 19\% shorter training time. Furthermore, PaCA reduces memory usage by 28\% and training time by 58\% compared to DoRA, while achieving comparable scores. One interesting observation is that the memory usage of PaCA increases by approximately 4GB when the rank is raised from 64 to 128, whereas the memory usage remains almost unchanged when increasing the rank from 8 to 16 in Section \ref{sec: task specific fine-tuning}. This is because a higher rank requires more optimizer state memory and activation memory for fine-tuning the partial connections. 

\subsection{QPaCA: Enhancements to QLoRA} \label{sec: qpac: enhancements to qlora}

\begin{table}[!htb]
\centering
\renewcommand{\arraystretch}{1.1} 
\caption{Comparisons of memory usage (Mem), training time (Time), and score on MT-Bench dataset when fine-tuning \textcolor{black}{LLaMA3-8B} and LLaMA3.1-70B on Oasst1 dataset using QLoRA and QPaCA. \textcolor{black}{No tuning and Quantized in the table refer to the models in 16-bit precision without quantization and with 4-bit NormalFloat Quantization (NF), respectively, without fine-tuning.}}
\label{tb: qpaca}
\resizebox{\textwidth}{!}{%
\begin{tabular}{c|c|cc|cccccccc|c}
\toprule
\textbf{Model} & \textbf{Method}  & \textbf{Mem} & \textbf{Time} & \textbf{Hums.} & \textbf{STEM} & \textbf{Role.} & \textbf{Extract.} & \textbf{Writing} & \textbf{Reason.} & \textbf{Coding} & \textbf{Math} & \textbf{Avg.} \\
\midrule
\multirow{4}{*}{\begin{tabular}[c]{@{}c@{}}\textbf{LLaMA3}\\ \textbf{-8B}\end{tabular}} & No tuning  & - & - & 6.25 & 5.70 & 5.45 & 4.85 & 5.20 & 4.40 & 3.20 & 1.95 & 4.62 \\ & Quantized & - & - & 4.70 & 4.80 & 4.60 & 5.00 & 4.65 & 4.05 & 3.60 & 1.85 & 4.16 \\& QLoRA & 18G & 42m & 6.85 & 5.75 & 5.85 & 6.00 & 5.15 & 4.70 & 3.35 & 2.35 & 5.00 \\ & \cellcolor{blue!10} QPaCA & \cellcolor{blue!10}\textbf{16G} & \cellcolor{blue!10}\textbf{37m} & \cellcolor{blue!10}6.85 & \cellcolor{blue!10}5.95 & \cellcolor{blue!10}5.65 & \cellcolor{blue!10}5.60 & \cellcolor{blue!10}5.15 & \cellcolor{blue!10}4.05 & \cellcolor{blue!10}3.65 & \cellcolor{blue!10}3.25 & \cellcolor{blue!10}\textbf{5.02} \\
\midrule
\multirow{3}{*}{\begin{tabular}[c]{@{}c@{}}\textbf{LLaMA3.1}\\ \textbf{-70B}\end{tabular}} & Quantized  & - & - & 7.40 & 7.05 & 5.85 & 6.50 & 6.85 & 5.30 & 4.60 & 3.80 & 5.92 \\ & QLoRA & 80G & 5.1h & 7.40 & 6.85 & 6.55 & 7.20 & 6.55 & 5.65 & 4.75 & 3.80 & \textbf{6.09} \\
 & \cellcolor{blue!10} QPaCA & \cellcolor{blue!10}\textbf{69G} & \cellcolor{blue!10}\textbf{4.7h} & \cellcolor{blue!10}7.70 & \cellcolor{blue!10}7.40 & \cellcolor{blue!10}6.40 & \cellcolor{blue!10}6.80 & \cellcolor{blue!10}6.50 & \cellcolor{blue!10}5.40 & \cellcolor{blue!10}4.75 & \cellcolor{blue!10}3.70 & \cellcolor{blue!10}6.08 \\
\bottomrule
\end{tabular}%
}
\end{table}

While PEFT significantly reduces the memory required for gradients and optimizer states, the model weights must be loaded onto the GPU, which consumes a significant amount of memory, especially when training large models. For example, loading the weights of LLaMA3.1-70B requires 140GB of memory, making it impossible to fine-tune using a single NVIDIA A100 GPU. To address this issue, QLoRA \citep{qlora-dettmers} quantizes the pretrained weights to 4 bits to further reduce memory usage and trains only the 16-bit adapter layers, enabling the fine-tuning of LLaMA3.1-70B on a single NVIDIA A100 GPU. This approach can be extended to PaCA by quantizing the unselected connections within the pretrained weights to 4 bits, while fine-tuning only the 16-bit randomly selected partial connections. We named this algorithm Quantized Partial Connection Adaptation (QPaCA) and compared it with QLoRA when fine-tuning \textcolor{black}{LLaMA3-8B} and LLaMA3.1-70B on the Oasst1 dataset using a single NVIDIA A100 GPU. Following Section \ref{sec: instruction tuning}, we evaluated the score on the MT-Bench dataset, using GPT4o-mini as the judge. Further details can be found in Table \ref{tb: hyperparameters, qlora} in Appendix \ref{appendix: experimental details}. 

Experimental results demonstrate that QPaCA reduces both memory usage and training time compared to QLoRA, as displayed in Table \ref{tb: qpaca}. Specifically, \textcolor{black}{on the LLaMA3-8B model, QPaCA not only achieved higher scores than the model quantized in the NF4 format, but also outperformed the 16-bit baseline, similar to QLoRA. Furthermore, QLoRA achieved an 11\% reduction in memory usage and a 12\% reduction in training time compared to QPaCA.}

\textcolor{black}{In addition, even on a larger scale model, LLaMA3.1-70B, QPaCA successfully reduces memory usage by 14\% and training time by 8\% with almost no drop in score compared to QLoRA and higher scores than the NF4 quantized model without fine-tuning on the MT-Bench dataset.} This training time reduction is relatively smaller than when comparing PaCA with LoRA in previous sections, and this is due to the time overheads of additional quantization and dequantization processes, which cannot be reduced by training only partial connections, unlike the forward and backward propagations.

\subsection{Usability of PaCA} \label{sec: training performance of paca}

\begin{table}[ht!]
\centering
\caption{Max sequence length for fine-tuning LLaMA3-8B using vaious PEFT algorithms on a single NVIDIA A100 GPU.}
\label{tb: maximum sequence length}
\begin{tabular}{c|cccc}
\toprule
\textbf{Method} & LoRA & DoRA & MosLoRA &  PaCA (Ours) \\ \midrule
\textbf{Max Length} & 8.0K & 4.7K & 8.0K &  \textbf{9.8K} \\ \bottomrule
\end{tabular}
\end{table}

In this section, we evaluate the usability of PaCA by measuring its training performance in different scenarios. We first increase the sequence length of the data while fine-tuning the LLaMA3-8B model with each PEFT method until an out-of-memory (OOM) error occurs, and the maximum sequence length is displayed in Table \ref{tb: maximum sequence length}. For a fair comparison, all other conditions, such as batch size and rank, were kept constant, except for the sequence length, as detailed in Table \ref{tb: hyperparameters, max length} in Appendix \ref{appendix: experimental details}. We found that PaCA increased the maximum sequence length by 23\%, 108\%, and 23\% compared to LoRA, DoRA, and MosLoRA, respectively, by storing only partial activations instead of all input activations.

\begin{figure}[!ht]
\centering

\subfloat[Throughput in a single A100 GPU. \label{fig: a100 throughput}]{\includegraphics[width=.47\textwidth]{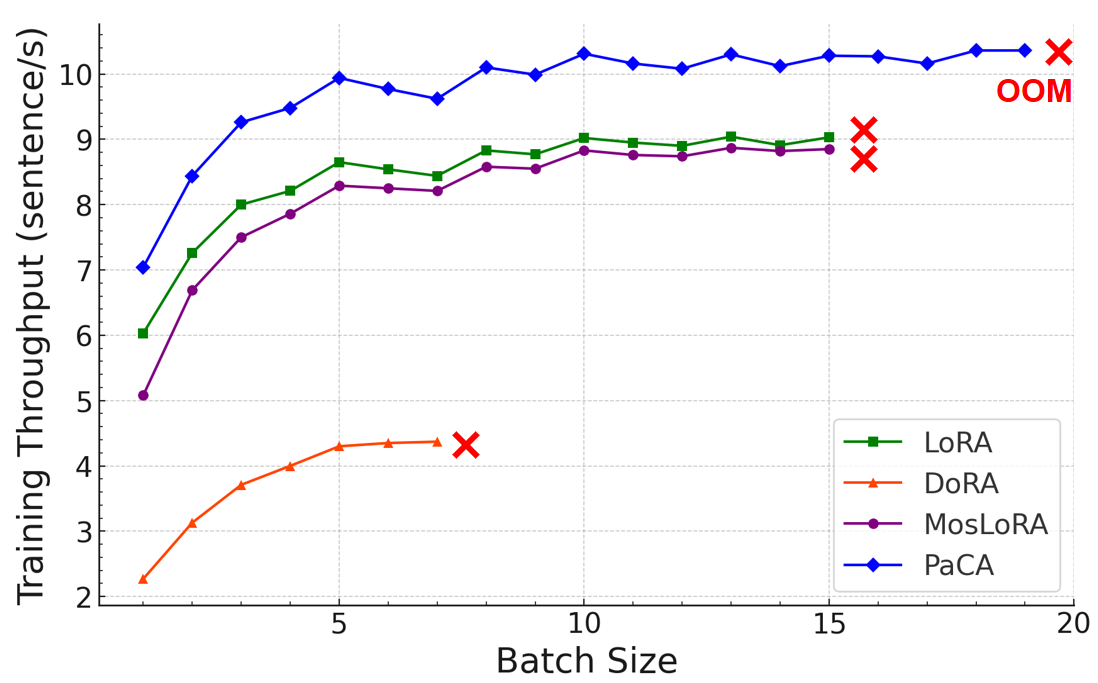}} \qquad
\subfloat[Throughput in a single Gaudi-v2 HPU. \label{fig: gaudi throughput.}]  {\includegraphics[width=.47\textwidth]{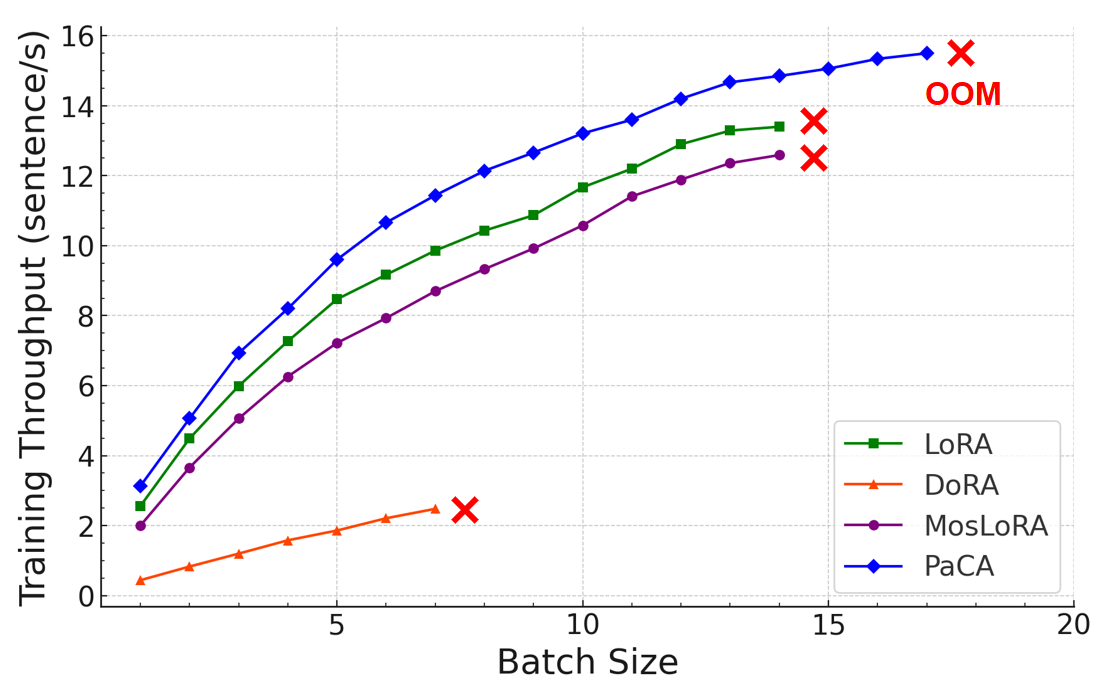}}
\caption{Training throughput (sentences/s) on a single NVIDIA A100 GPU and INTEL Gaudi2 HPU when fine-tuning LLaMA3-8B with a sequence length of 512.}
\label{fig: throughput of lora variants and paca}
\end{figure}

Next, we evaluate the training throughput improvements achieved by PaCA compared to LoRA and its variants as the batch size increases when fine-tuning LLaMA3-8B using a single NVIDIA A100 GPU and Intel Gaudi2 HPU. Specifically, we kept all configurations identical except for the batch size as presented in Table \ref{tb: hyperparameters, training throughput}  in Appendix \ref{appendix: experimental details}, and measured the throughput as the batch size increased for each PEFT method until an OOM error occurred. As shown in Fig. \ref{fig: throughput of lora variants and paca}, PaCA demonstrated the ability to increase the batch size by 33\% on the NVIDIA A100 GPU and 21\% on the Intel Gaudi2 HPU compared to LoRA and its variants, primarily due to its reduction of activation memory. This reduction allows PaCA to handle larger batch sizes, which directly leads to better resource utilization and improves scalability. In addition, at the same batch size, PaCA consistently achieved higher training throughput compared to LoRA and its variants, as PaCA eliminates inefficient sequential processing introduced by adapter layers, allowing for higher hardware utilization. Consequently, PaCA outperformed LoRA, achieving a throughput of 10.36 sentences/s on A100 GPU and 15.5 sentences/s on Gaudi2 HPU, representing a 16\% improvement for both GPUs. 

\section{\textcolor{black}{Effect of Selection Strategy}} \label{sec: ablation study}

\textcolor{black}{
In this section, we explore alternative strategies for selecting connections in PaCA and evaluate their effectiveness beyond the random selection approach. We tested two selection schemes that consider the importance of each column. A weight-based strategy selects the columns with the highest $L_2$-Norm from the initial pretrained weights, whereas a gradient-based strategy accumulates gradients during the first 100 iterations without updating weights (i.e., $G_i = \sum_t \|g_i^t\|^2$, where $i$ is the number of layers and $t$ is the accumulation step) and selects columns with the largest accumulated gradients. Experimental results are displayed in the table below.
}

\begin{table}[!htb]
\caption{\textcolor{black}{Test score on MT-Bench dataset when fine-tuning LLaMA3-8B with PaCA using various connection selecting strategy on Oasst1 dataset.}}
\label{tb: selecting strategy}
\centering
\renewcommand{\arraystretch}{1.1}
\resizebox{\textwidth}{!}{%
\begin{tabular}{c|cccccccc|c}
\toprule
\textbf{Method} & \textbf{Human.} & \textbf{STEM} & \textbf{Role.} & \textbf{Extract.} & \textbf{Writing} & \textbf{Reason.} & \textbf{Coding} & \textbf{Math} & \textbf{Avg.} \\ 
\midrule
No tuning & 6.25 & 5.70 & 5.45 & 4.85 & 5.20 & 4.40 & 3.20 & 1.95 & 4.62 \\
Random (Seed \#1) & 6.50 & 6.30 & 5.90 & 5.95 & 5.65 & 4.8 & 3.7 & 3.05 & 5.23 \\ 
Random (Seed \#2) & 6.50 & 6.00 & 6.30 & 5.90 & 5.70 & 4.90 & 3.80 & 3.00 & \textbf{5.26} \\ 
Weight-based & 7.00 & 5.70 & 6.05 & 5.80 & 5.70 & 4.55 & 3.90 & 2.70 & 5.18 \\ 
Gradient-based & 6.95 & 6.40 & 6.25 & 5.35 & 5.95 & 4.55 & 3.80 & 2.70 & 5.24 \\
\bottomrule
\end{tabular}%
}
\end{table}
\textcolor{black}{
Table \ref{tb: selecting strategy} demonstrates that random selection achieves similar performance to importance-based selection schemes. In other words, the choice of selection strategy does not noticeably affect fine-tuning accuracy. Therefore, we chose to select connections randomly in PaCA, as this strategy eliminates the need for complex processes to measure the importance of connections, thereby minimizing training time or memory overhead without performance degradation.}

\section{Related Work} \label{sec: related works}
\paragraph{Parameter-efficient fine-tuning (PEFT)}
Fine-tuning LLMs requires significant memory resources to store parameter gradients and optimizer states. PEFT algorithms address this challenge by introducing adapter layers with far fewer parameters than the pretrained models, significantly reducing the memory required for parameter gradients and optimizer states by fine-tuning only the adapter layers. PEFT methods can generally be categorized into three groups: \textit{Adapter-based methods} \citep{prefix-Lisa, series-adapter-Houlsby, parallel-adapter-He}, \textit{Prompt-based methods} \citep{prompt-tuning-Lester, Residual-Prompt-Tuning-Razdaibiedina, Input-Prompt-Wang, LLaMA-Adapter-Zhang, LLaMA-Adapter-v2-Gao}, and \textit{LoRA and its variants} \citep{LoRA-Hu, Vera-Kopiczko, DoRA-Yang, MosLoRA-Wu}. \textit{Adapter-based methods} introduce new trainable adapter weights to the pretrained models. For example, \citet{series-adapter-Houlsby} adds adapter layers as linear modules in series with the existing model, while \citet{parallel-adapter-He} inserts adapter modules in parallel with the pretrained model. Secondly, \textit{Prompt-based methods} inject new trainable prompt vectors into the model. Specifically, LLaMA-Adapter \citep{LLaMA-Adapter-Zhang, LLaMA-Adapter-v2-Gao} introduces prompts into the upper layers of the transformer, enabling the model to incorporate diverse knowledge. Although these approaches enable efficient fine-tuning with smaller trainable parameters, they introduce latency overhead during inference due to the sequential processing of the adapter layers and the pretrained model.

The third category of PEFT methods is \textit{LoRA and its variants}. LoRA \citep{LoRA-Hu} introduces low-rank matrices as adapters to approximate weight gradients during fine-tuning, then merges these low-rank matrices with the pretrained weights, effectively eliminating inference overhead. VeRA \citep{Vera-Kopiczko} takes this approach further by freezing the low-rank matrices and sharing them across layers, while only learning the scaling vectors for each layer, which significantly reduces the number of trainable parameters. DoRA \citep{DoRA-Yang} improves upon LoRA by considering both the magnitude and direction of gradients through weight decomposition, leading to higher accuracy compared to LoRA. MosLoRA \citep{MosLoRA-Wu} enhances LoRA by introducing a learnable mixer between the two low-rank matrices, improving its capabilities. SHiRA \citep{sharp-kartikeya} fine-tunes a sparse 1–2\% subset of the pretrained weights, thereby enabling rapid adapter switching during inference in mobile environments. Even though \textit{LoRA and its variants} can remove latency overhead by merging the adapter weights with the pretrained weights during inference, latency overhead persists during fine-tuning, as merging the weights is not feasible at this stage.

\paragraph{PEFT with Quantization}

Quantization \citep{LLM-int8-Dettmers, GPTQ-Frantar, AWQ-Lin, OWQ-Frantar} is a technique that reduces memory usage and computational complexity by representing weights or activations in low precision. This method can also be combined with PEFT to reduce memory usage during fine-tuning \citep{Alpha-tuning-kwon, qlora-dettmers, Xu-QALoRA}. For instance, QLoRA \citep{qlora-dettmers} compresses pretrained weights to 4 bits and trains only the low-rank adapter matrices represented in 16 bits, significantly reducing the memory required to load the model. Additionally, QA-LoRA \citep{Xu-QALoRA} integrates the low-rank adapter matrices with the zero point in quantization, enabling the direct generation of a 4-bit quantized model after fine-tuning. While those quantized-PEFT approaches reduce memory usage for fine-tuning, the sequential processes introduced by the adapter layers still cause training time overhead.

\section{Conclusion}

In this work, we propose PaCA, a memory-efficient PEFT algorithm that fine-tunes randomly selected partial connections within the pretrained weights without employing additional adapter layers. By removing the sequential processing overhead associated with the adapters in prior PEFT schemes, PaCA significantly improves hardware utilization and training speed. In addition, PaCA reduces activation memory by only storing partial activations instead of all input activations. We theoretically prove that PaCA can successfully converge in general deep neural networks. Moreover, in experiments, PaCA consistently outperforms LoRA and its variants in training performance while maintaining comparable accuracy across various fine-tuning scenarios. We also show that PaCA can be applied simultaneously with quantization. Finally, we demonstrate the effectiveness of PaCA in scenarios involving long sequence data or when maximizing throughput in resource-constrained environments. For future work, we aim to develop methods for identifying optimal partial connections in PaCA, rather than relying on random selection, to further enhance fine-tuning accuracy.

\section*{Reproducibility}

We introduce PaCA and provide a detailed explanation of its concept and potential in Section \ref{sec: paca: partial connection adaptation}, and prove its theoretical convergence in Section \ref{sec: convergence analysis of paca}. In addition, the setup and hyperparameters are thoroughly explained in Section \ref{sec: experiments} and Appendix \ref{appendix: experimental details}. Furthermore, we have implemented PaCA using PyTorch \citep{pytorch}, a widely used deep learning framework, and integrated it into the PEFT library in Huggingface \citep{huggingface} to ensure easy reproducibility.

\section*{Acknowledgment}

This work was supported by the National Research Foundation of Korea (Grant NRF-
2022R1C1C1006880), the Institute of Information \& Communications Technology Planning
\& Evaluation (Grant IITP-2023-RS-2023-00256081 and Grant RS-2024-00347394), and the NAVER-Intel Co-Lab.


\bibliography{ref}

\begin{thebibliography}{54}
\providecommand{\natexlab}[1]{#1}
\providecommand{\url}[1]{\texttt{#1}}
\expandafter\ifx\csname urlstyle\endcsname\relax
  \providecommand{\doi}[1]{doi: #1}\else
  \providecommand{\doi}{doi: \begingroup \urlstyle{rm}\Url}\fi

\bibitem[Aghajanyan et~al.(2021)Aghajanyan, Gupta, and Zettlemoyer]{Aghajanyan-subspace2}
Armen Aghajanyan, Sonal Gupta, and Luke Zettlemoyer.
\newblock Intrinsic dimensionality explains the effectiveness of language model fine-tuning.
\newblock In Chengqing Zong, Fei Xia, Wenjie Li, and Roberto Navigli (eds.), \emph{Proceedings of the 59th Annual Meeting of the Association for Computational Linguistics and the 11th International Joint Conference on Natural Language Processing, {ACL/IJCNLP} 2021, (Volume 1: Long Papers), Virtual Event, August 1-6, 2021}, pp.\  7319--7328. Association for Computational Linguistics, 2021.
\newblock \doi{10.18653/V1/2021.ACL-LONG.568}.
\newblock URL \url{https://doi.org/10.18653/v1/2021.acl-long.568}.

\bibitem[Belilovsky et~al.(2020)Belilovsky, Eickenberg, and Oyallon]{DGL-Beliovsky}
Eugene Belilovsky, Michael Eickenberg, and Edouard Oyallon.
\newblock Decoupled greedy learning of cnns.
\newblock In \emph{Proceedings of the 37th International Conference on Machine Learning, {ICML} 2020, 13-18 July 2020, Virtual Event}, volume 119 of \emph{Proceedings of Machine Learning Research}, pp.\  736--745. {PMLR}, 2020.
\newblock URL \url{http://proceedings.mlr.press/v119/belilovsky20a.html}.

\bibitem[Bhardwaj et~al.(2024)Bhardwaj, Pandey, Priyadarshi, Ganapathy, Kadambi, Esteves, Borse, Whatmough, Garrepalli, Baalen, Teague, and Nagel]{sharp-kartikeya}
Kartikeya Bhardwaj, Nilesh~Prasad Pandey, Sweta Priyadarshi, Viswanath Ganapathy, Shreya Kadambi, Rafael Esteves, Shubhankar Borse, Paul Whatmough, Risheek Garrepalli, Mart~Van Baalen, Harris Teague, and Markus Nagel.
\newblock Sparse high rank adapters.
\newblock In \emph{The Thirty-eighth Annual Conference on Neural Information Processing Systems}, 2024.
\newblock URL \url{https://openreview.net/forum?id=6hY60tkiEK}.

\bibitem[Chen et~al.(2021)Chen, Zheng, Yao, Wang, Stoica, Mahoney, and Gonzalez]{ActNN-Chen}
Jianfei Chen, Lianmin Zheng, Zhewei Yao, Dequan Wang, Ion Stoica, Michael~W. Mahoney, and Joseph Gonzalez.
\newblock Actnn: Reducing training memory footprint via 2-bit activation compressed training.
\newblock In Marina Meila and Tong Zhang (eds.), \emph{Proceedings of the 38th International Conference on Machine Learning, {ICML} 2021, 18-24 July 2021, Virtual Event}, volume 139 of \emph{Proceedings of Machine Learning Research}, pp.\  1803--1813. {PMLR}, 2021.
\newblock URL \url{http://proceedings.mlr.press/v139/chen21z.html}.

\bibitem[Chen et~al.(2023)Chen, Xu, Wang, Cheng, and Yao]{DropIT-Chen}
Joya Chen, Kai Xu, Yuhui Wang, Yifei Cheng, and Angela Yao.
\newblock Dropit: Dropping intermediate tensors for memory-efficient {DNN} training.
\newblock In \emph{The Eleventh International Conference on Learning Representations, {ICLR} 2023, Kigali, Rwanda, May 1-5, 2023}. OpenReview.net, 2023.
\newblock URL \url{https://openreview.net/forum?id=Kn6i2BZW69w}.

\bibitem[Choquette et~al.(2021)Choquette, Gandhi, Giroux, Stam, and Krashinsky]{a100-choquette}
Jack Choquette, Wishwesh Gandhi, Olivier Giroux, Nick Stam, and Ronny Krashinsky.
\newblock {NVIDIA} {A100} tensor core {GPU:} performance and innovation.
\newblock \emph{{IEEE} Micro}, 41\penalty0 (2):\penalty0 29--35, 2021.
\newblock \doi{10.1109/MM.2021.3061394}.
\newblock URL \url{https://doi.org/10.1109/MM.2021.3061394}.

\bibitem[Dai et~al.(2018)Dai, Lin, Li, Zhao, Wang, Zheng, and Zhou]{dai-multikernel2}
Hongwen Dai, Zhen Lin, Chao Li, Chen Zhao, Fei Wang, Nanning Zheng, and Huiyang Zhou.
\newblock Accelerate gpu concurrent kernel execution by mitigating memory pipeline stalls.
\newblock In \emph{2018 IEEE international symposium on high performance computer architecture (HPCA)}, pp.\  208--220. IEEE, 2018.

\bibitem[Dettmers et~al.(2022)Dettmers, Lewis, Belkada, and Zettlemoyer]{LLM-int8-Dettmers}
Tim Dettmers, Mike Lewis, Younes Belkada, and Luke Zettlemoyer.
\newblock Llm.int8(): 8-bit matrix multiplication for transformers at scale.
\newblock \emph{CoRR}, abs/2208.07339, 2022.
\newblock \doi{10.48550/ARXIV.2208.07339}.
\newblock URL \url{https://doi.org/10.48550/arXiv.2208.07339}.

\bibitem[Dettmers et~al.(2023)Dettmers, Pagnoni, Holtzman, and Zettlemoyer]{qlora-dettmers}
Tim Dettmers, Artidoro Pagnoni, Ari Holtzman, and Luke Zettlemoyer.
\newblock Qlora: Efficient finetuning of quantized llms.
\newblock In Alice Oh, Tristan Naumann, Amir Globerson, Kate Saenko, Moritz Hardt, and Sergey Levine (eds.), \emph{Advances in Neural Information Processing Systems 36: Annual Conference on Neural Information Processing Systems 2023, NeurIPS 2023, New Orleans, LA, USA, December 10 - 16, 2023}, 2023.
\newblock URL \url{http://papers.nips.cc/paper\_files/paper/2023/hash/1feb87871436031bdc0f2beaa62a049b-Abstract-Conference.html}.

\bibitem[Dosovitskiy et~al.(2021)Dosovitskiy, Beyer, Kolesnikov, Weissenborn, Zhai, Unterthiner, Dehghani, Minderer, Heigold, Gelly, Uszkoreit, and Houlsby]{vit-dosovitskiy}
Alexey Dosovitskiy, Lucas Beyer, Alexander Kolesnikov, Dirk Weissenborn, Xiaohua Zhai, Thomas Unterthiner, Mostafa Dehghani, Matthias Minderer, Georg Heigold, Sylvain Gelly, Jakob Uszkoreit, and Neil Houlsby.
\newblock An image is worth 16x16 words: Transformers for image recognition at scale.
\newblock In \emph{9th International Conference on Learning Representations, {ICLR} 2021, Virtual Event, Austria, May 3-7, 2021}. OpenReview.net, 2021.
\newblock URL \url{https://openreview.net/forum?id=YicbFdNTTy}.

\bibitem[Dubey et~al.(2024)Dubey, Jauhri, Pandey, Kadian, Al{-}Dahle, Letman, Mathur, Schelten, Yang, Fan, Goyal, Hartshorn, Yang, Mitra, Sravankumar, Korenev, Hinsvark, Rao, Zhang, Rodriguez, Gregerson, Spataru, Rozi{\`{e}}re, Biron, Tang, Chern, Caucheteux, Nayak, Bi, Marra, McConnell, Keller, Touret, Wu, Wong, Ferrer, Nikolaidis, Allonsius, Song, Pintz, Livshits, Esiobu, Choudhary, Mahajan, Garcia{-}Olano, Perino, Hupkes, Lakomkin, AlBadawy, Lobanova, Dinan, Smith, Radenovic, Zhang, Synnaeve, Lee, Anderson, Nail, Mialon, Pang, Cucurell, Nguyen, Korevaar, Xu, Touvron, Zarov, Ibarra, Kloumann, Misra, Evtimov, Copet, Lee, Geffert, Vranes, Park, Mahadeokar, Shah, van~der Linde, Billock, Hong, Lee, Fu, Chi, Huang, Liu, Wang, Yu, Bitton, Spisak, Park, Rocca, Johnstun, Saxe, Jia, Alwala, Upasani, Plawiak, Li, Heafield, Stone, and et~al.]{LLaMA3-Dubey}
Abhimanyu Dubey, Abhinav Jauhri, Abhinav Pandey, Abhishek Kadian, Ahmad Al{-}Dahle, Aiesha Letman, Akhil Mathur, Alan Schelten, Amy Yang, Angela Fan, Anirudh Goyal, Anthony Hartshorn, Aobo Yang, Archi Mitra, Archie Sravankumar, Artem Korenev, Arthur Hinsvark, Arun Rao, Aston Zhang, Aur{\'{e}}lien Rodriguez, Austen Gregerson, Ava Spataru, Baptiste Rozi{\`{e}}re, Bethany Biron, Binh Tang, Bobbie Chern, Charlotte Caucheteux, Chaya Nayak, Chloe Bi, Chris Marra, Chris McConnell, Christian Keller, Christophe Touret, Chunyang Wu, Corinne Wong, Cristian~Canton Ferrer, Cyrus Nikolaidis, Damien Allonsius, Daniel Song, Danielle Pintz, Danny Livshits, David Esiobu, Dhruv Choudhary, Dhruv Mahajan, Diego Garcia{-}Olano, Diego Perino, Dieuwke Hupkes, Egor Lakomkin, Ehab AlBadawy, Elina Lobanova, Emily Dinan, Eric~Michael Smith, Filip Radenovic, Frank Zhang, Gabriel Synnaeve, Gabrielle Lee, Georgia~Lewis Anderson, Graeme Nail, Gr{\'{e}}goire Mialon, Guan Pang, Guillem Cucurell, Hailey Nguyen, Hannah Korevaar, Hu~Xu, Hugo
  Touvron, Iliyan Zarov, Imanol~Arrieta Ibarra, Isabel~M. Kloumann, Ishan Misra, Ivan Evtimov, Jade Copet, Jaewon Lee, Jan Geffert, Jana Vranes, Jason Park, Jay Mahadeokar, Jeet Shah, Jelmer van~der Linde, Jennifer Billock, Jenny Hong, Jenya Lee, Jeremy Fu, Jianfeng Chi, Jianyu Huang, Jiawen Liu, Jie Wang, Jiecao Yu, Joanna Bitton, Joe Spisak, Jongsoo Park, Joseph Rocca, Joshua Johnstun, Joshua Saxe, Junteng Jia, Kalyan~Vasuden Alwala, Kartikeya Upasani, Kate Plawiak, Ke~Li, Kenneth Heafield, Kevin Stone, and et~al.
\newblock The llama 3 herd of models.
\newblock \emph{CoRR}, abs/2407.21783, 2024.
\newblock \doi{10.48550/ARXIV.2407.21783}.
\newblock URL \url{https://doi.org/10.48550/arXiv.2407.21783}.

\bibitem[Frantar et~al.(2022)Frantar, Ashkboos, Hoefler, and Alistarh]{GPTQ-Frantar}
Elias Frantar, Saleh Ashkboos, Torsten Hoefler, and Dan Alistarh.
\newblock {GPTQ:} accurate post-training quantization for generative pre-trained transformers.
\newblock \emph{CoRR}, abs/2210.17323, 2022.
\newblock \doi{10.48550/ARXIV.2210.17323}.
\newblock URL \url{https://doi.org/10.48550/arXiv.2210.17323}.

\bibitem[Frantar et~al.(2023)Frantar, Ashkboos, Hoefler, and Alistarh]{OWQ-Frantar}
Elias Frantar, Saleh Ashkboos, Torsten Hoefler, and Dan Alistarh.
\newblock {OPTQ:} accurate quantization for generative pre-trained transformers.
\newblock In \emph{The Eleventh International Conference on Learning Representations, {ICLR} 2023, Kigali, Rwanda, May 1-5, 2023}. OpenReview.net, 2023.
\newblock URL \url{https://openreview.net/forum?id=tcbBPnfwxS}.

\bibitem[Gao et~al.(2023)Gao, Han, Zhang, Lin, Geng, Zhou, Zhang, Lu, He, Yue, Li, and Qiao]{LLaMA-Adapter-v2-Gao}
Peng Gao, Jiaming Han, Renrui Zhang, Ziyi Lin, Shijie Geng, Aojun Zhou, Wei Zhang, Pan Lu, Conghui He, Xiangyu Yue, Hongsheng Li, and Yu~Qiao.
\newblock Llama-adapter {V2:} parameter-efficient visual instruction model.
\newblock \emph{CoRR}, abs/2304.15010, 2023.
\newblock \doi{10.48550/ARXIV.2304.15010}.
\newblock URL \url{https://doi.org/10.48550/arXiv.2304.15010}.

\bibitem[Han et~al.(2022)Han, Zhang, Chen, and Chen]{han-multikernel3}
Mingcong Han, Hanze Zhang, Rong Chen, and Haibo Chen.
\newblock Microsecond-scale preemption for concurrent $\{$GPU-accelerated$\}$$\{$DNN$\}$ inferences.
\newblock In \emph{16th USENIX Symposium on Operating Systems Design and Implementation (OSDI 22)}, pp.\  539--558, 2022.

\bibitem[He et~al.(2022)He, Zhou, Ma, Berg{-}Kirkpatrick, and Neubig]{parallel-adapter-He}
Junxian He, Chunting Zhou, Xuezhe Ma, Taylor Berg{-}Kirkpatrick, and Graham Neubig.
\newblock Towards a unified view of parameter-efficient transfer learning.
\newblock In \emph{The Tenth International Conference on Learning Representations, {ICLR} 2022, Virtual Event, April 25-29, 2022}. OpenReview.net, 2022.
\newblock URL \url{https://openreview.net/forum?id=0RDcd5Axok}.

\bibitem[Hendrycks et~al.(2021)Hendrycks, Burns, Basart, Zou, Mazeika, Song, and Steinhardt]{mmlu-hendryckstest2021}
Dan Hendrycks, Collin Burns, Steven Basart, Andy Zou, Mantas Mazeika, Dawn Song, and Jacob Steinhardt.
\newblock Measuring massive multitask language understanding.
\newblock \emph{Proceedings of the International Conference on Learning Representations (ICLR)}, 2021.

\bibitem[Hoffmann et~al.(2022)Hoffmann, Borgeaud, Mensch, Buchatskaya, Cai, Rutherford, de~Las~Casas, Hendricks, Welbl, Clark, Hennigan, Noland, Millican, van~den Driessche, Damoc, Guy, Osindero, Simonyan, Elsen, Rae, Vinyals, and Sifre]{Hoffmann-scaling-laws2}
Jordan Hoffmann, Sebastian Borgeaud, Arthur Mensch, Elena Buchatskaya, Trevor Cai, Eliza Rutherford, Diego de~Las~Casas, Lisa~Anne Hendricks, Johannes Welbl, Aidan Clark, Tom Hennigan, Eric Noland, Katie Millican, George van~den Driessche, Bogdan Damoc, Aurelia Guy, Simon Osindero, Karen Simonyan, Erich Elsen, Jack~W. Rae, Oriol Vinyals, and Laurent Sifre.
\newblock Training compute-optimal large language models.
\newblock \emph{CoRR}, abs/2203.15556, 2022.
\newblock \doi{10.48550/ARXIV.2203.15556}.
\newblock URL \url{https://doi.org/10.48550/arXiv.2203.15556}.

\bibitem[Houlsby et~al.(2019)Houlsby, Giurgiu, Jastrzebski, Morrone, de~Laroussilhe, Gesmundo, Attariyan, and Gelly]{series-adapter-Houlsby}
Neil Houlsby, Andrei Giurgiu, Stanislaw Jastrzebski, Bruna Morrone, Quentin de~Laroussilhe, Andrea Gesmundo, Mona Attariyan, and Sylvain Gelly.
\newblock Parameter-efficient transfer learning for {NLP}.
\newblock In Kamalika Chaudhuri and Ruslan Salakhutdinov (eds.), \emph{Proceedings of the 36th International Conference on Machine Learning, {ICML} 2019, 9-15 June 2019, Long Beach, California, {USA}}, volume~97 of \emph{Proceedings of Machine Learning Research}, pp.\  2790--2799. {PMLR}, 2019.
\newblock URL \url{http://proceedings.mlr.press/v97/houlsby19a.html}.

\bibitem[Hu et~al.(2022)Hu, Shen, Wallis, Allen{-}Zhu, Li, Wang, Wang, and Chen]{LoRA-Hu}
Edward~J. Hu, Yelong Shen, Phillip Wallis, Zeyuan Allen{-}Zhu, Yuanzhi Li, Shean Wang, Lu~Wang, and Weizhu Chen.
\newblock Lora: Low-rank adaptation of large language models.
\newblock In \emph{The Tenth International Conference on Learning Representations, {ICLR} 2022, Virtual Event, April 25-29, 2022}. OpenReview.net, 2022.
\newblock URL \url{https://openreview.net/forum?id=nZeVKeeFYf9}.

\bibitem[{Intel Corporation}(2023)]{gaudi2-interl}
{Intel Corporation}.
\newblock Intel gaudi2 ai accelerators white paper.
\newblock Technical report, Intel Corporation, 2023.
\newblock Accessed: 2024-09-28.

\bibitem[Kaplan et~al.(2020)Kaplan, McCandlish, Henighan, Brown, Chess, Child, Gray, Radford, Wu, and Amodei]{Kaplan-scaling-laws1}
Jared Kaplan, Sam McCandlish, Tom Henighan, Tom~B. Brown, Benjamin Chess, Rewon Child, Scott Gray, Alec Radford, Jeffrey Wu, and Dario Amodei.
\newblock Scaling laws for neural language models.
\newblock \emph{CoRR}, abs/2001.08361, 2020.
\newblock URL \url{https://arxiv.org/abs/2001.08361}.

\bibitem[K{\"{o}}pf et~al.(2023)K{\"{o}}pf, Kilcher, von R{\"{u}}tte, Anagnostidis, Tam, Stevens, Barhoum, Nguyen, Stanley, Nagyfi, ES, Suri, Glushkov, Dantuluri, Maguire, Schuhmann, Nguyen, and Mattick]{oasst1-kopf}
Andreas K{\"{o}}pf, Yannic Kilcher, Dimitri von R{\"{u}}tte, Sotiris Anagnostidis, Zhi~Rui Tam, Keith Stevens, Abdullah Barhoum, Duc Nguyen, Oliver Stanley, Rich{\'{a}}rd Nagyfi, Shahul ES, Sameer Suri, David Glushkov, Arnav Dantuluri, Andrew Maguire, Christoph Schuhmann, Huu Nguyen, and Alexander Mattick.
\newblock Openassistant conversations - democratizing large language model alignment.
\newblock In Alice Oh, Tristan Naumann, Amir Globerson, Kate Saenko, Moritz Hardt, and Sergey Levine (eds.), \emph{Advances in Neural Information Processing Systems 36: Annual Conference on Neural Information Processing Systems 2023, NeurIPS 2023, New Orleans, LA, USA, December 10 - 16, 2023}, 2023.
\newblock URL \url{http://papers.nips.cc/paper\_files/paper/2023/hash/949f0f8f32267d297c2d4e3ee10a2e7e-Abstract-Datasets\_and\_Benchmarks.html}.

\bibitem[Kopiczko et~al.(2024)Kopiczko, Blankevoort, and Asano]{Vera-Kopiczko}
Dawid~Jan Kopiczko, Tijmen Blankevoort, and Yuki~M. Asano.
\newblock Vera: Vector-based random matrix adaptation.
\newblock In \emph{The Twelfth International Conference on Learning Representations, {ICLR} 2024, Vienna, Austria, May 7-11, 2024}. OpenReview.net, 2024.
\newblock URL \url{https://openreview.net/forum?id=NjNfLdxr3A}.

\bibitem[Korthikanti et~al.(2023)Korthikanti, Casper, Lym, McAfee, Andersch, Shoeybi, and Catanzaro]{Megatron-Korthikanti}
Vijay~Anand Korthikanti, Jared Casper, Sangkug Lym, Lawrence McAfee, Michael Andersch, Mohammad Shoeybi, and Bryan Catanzaro.
\newblock Reducing activation recomputation in large transformer models.
\newblock In Dawn Song, Michael Carbin, and Tianqi Chen (eds.), \emph{Proceedings of the Sixth Conference on Machine Learning and Systems, MLSys 2023, Miami, FL, USA, June 4-8, 2023}. mlsys.org, 2023.
\newblock URL \url{https://proceedings.mlsys.org/paper\_files/paper/2023/hash/80083951326cf5b35e5100260d64ed81-Abstract-mlsys2023.html}.

\bibitem[Krizhevsky \& Hinton(2009)Krizhevsky and Hinton]{cifar10-krizhevsky}
Alex Krizhevsky and Geoffrey Hinton.
\newblock Learning multiple layers of features from tiny images.
\newblock Technical report, University of Toronto, 2009.

\bibitem[Kwon et~al.(2022)Kwon, Kim, Bae, Yoo, Kim, Park, Kim, Ha, Sung, and Lee]{Alpha-tuning-kwon}
Se~Jung Kwon, Jeonghoon Kim, Jeongin Bae, Kang~Min Yoo, Jin{-}Hwa Kim, Baeseong Park, Byeongwook Kim, Jung{-}Woo Ha, Nako Sung, and Dongsoo Lee.
\newblock Alphatuning: Quantization-aware parameter-efficient adaptation of large-scale pre-trained language models.
\newblock In Yoav Goldberg, Zornitsa Kozareva, and Yue Zhang (eds.), \emph{Findings of the Association for Computational Linguistics: {EMNLP} 2022, Abu Dhabi, United Arab Emirates, December 7-11, 2022}, pp.\  3288--3305. Association for Computational Linguistics, 2022.
\newblock \doi{10.18653/V1/2022.FINDINGS-EMNLP.240}.
\newblock URL \url{https://doi.org/10.18653/v1/2022.findings-emnlp.240}.

\bibitem[Lester et~al.(2021)Lester, Al{-}Rfou, and Constant]{prompt-tuning-Lester}
Brian Lester, Rami Al{-}Rfou, and Noah Constant.
\newblock The power of scale for parameter-efficient prompt tuning.
\newblock In Marie{-}Francine Moens, Xuanjing Huang, Lucia Specia, and Scott~Wen{-}tau Yih (eds.), \emph{Proceedings of the 2021 Conference on Empirical Methods in Natural Language Processing, {EMNLP} 2021, Virtual Event / Punta Cana, Dominican Republic, 7-11 November, 2021}, pp.\  3045--3059. Association for Computational Linguistics, 2021.
\newblock \doi{10.18653/V1/2021.EMNLP-MAIN.243}.
\newblock URL \url{https://doi.org/10.18653/v1/2021.emnlp-main.243}.

\bibitem[Li et~al.(2018)Li, Farkhoor, Liu, and Yosinski]{Li-subspace1}
Chunyuan Li, Heerad Farkhoor, Rosanne Liu, and Jason Yosinski.
\newblock Measuring the intrinsic dimension of objective landscapes.
\newblock In \emph{6th International Conference on Learning Representations, {ICLR} 2018, Vancouver, BC, Canada, April 30 - May 3, 2018, Conference Track Proceedings}. OpenReview.net, 2018.
\newblock URL \url{https://openreview.net/forum?id=ryup8-WCW}.

\bibitem[Li \& Liang(2021)Li and Liang]{prefix-Lisa}
Xiang~Lisa Li and Percy Liang.
\newblock Prefix-tuning: Optimizing continuous prompts for generation.
\newblock In Chengqing Zong, Fei Xia, Wenjie Li, and Roberto Navigli (eds.), \emph{Proceedings of the 59th Annual Meeting of the Association for Computational Linguistics and the 11th International Joint Conference on Natural Language Processing, {ACL/IJCNLP} 2021, (Volume 1: Long Papers), Virtual Event, August 1-6, 2021}, pp.\  4582--4597. Association for Computational Linguistics, 2021.
\newblock \doi{10.18653/V1/2021.ACL-LONG.353}.
\newblock URL \url{https://doi.org/10.18653/v1/2021.acl-long.353}.

\bibitem[Lin et~al.(2024)Lin, Tang, Tang, Yang, Chen, Wang, Xiao, Dang, Gan, and Han]{AWQ-Lin}
Ji~Lin, Jiaming Tang, Haotian Tang, Shang Yang, Wei{-}Ming Chen, Wei{-}Chen Wang, Guangxuan Xiao, Xingyu Dang, Chuang Gan, and Song Han.
\newblock {AWQ:} activation-aware weight quantization for on-device {LLM} compression and acceleration.
\newblock In Phillip~B. Gibbons, Gennady Pekhimenko, and Christopher~De Sa (eds.), \emph{Proceedings of the Seventh Annual Conference on Machine Learning and Systems, MLSys 2024, Santa Clara, CA, USA, May 13-16, 2024}. mlsys.org, 2024.
\newblock URL \url{https://proceedings.mlsys.org/paper\_files/paper/2024/hash/42a452cbafa9dd64e9ba4aa95cc1ef21-Abstract-Conference.html}.

\bibitem[Liu et~al.(2024)Liu, Wang, Yin, Molchanov, Wang, Cheng, and Chen]{DoRA-Yang}
Shih{-}Yang Liu, Chien{-}Yi Wang, Hongxu Yin, Pavlo Molchanov, Yu{-}Chiang~Frank Wang, Kwang{-}Ting Cheng, and Min{-}Hung Chen.
\newblock Dora: Weight-decomposed low-rank adaptation.
\newblock In \emph{Forty-first International Conference on Machine Learning, {ICML} 2024, Vienna, Austria, July 21-27, 2024}. OpenReview.net, 2024.
\newblock URL \url{https://openreview.net/forum?id=3d5CIRG1n2}.

\bibitem[Liu et~al.(2022)Liu, Zheng, Wang, Cen, Chen, Han, Chen, Liu, Tang, Gonzalez, Mahoney, and Cheung]{GACT-Liu}
Xiaoxuan Liu, Lianmin Zheng, Dequan Wang, Yukuo Cen, Weize Chen, Xu~Han, Jianfei Chen, Zhiyuan Liu, Jie Tang, Joey Gonzalez, Michael~W. Mahoney, and Alvin Cheung.
\newblock {GACT:} activation compressed training for generic network architectures.
\newblock In Kamalika Chaudhuri, Stefanie Jegelka, Le~Song, Csaba Szepesv{\'{a}}ri, Gang Niu, and Sivan Sabato (eds.), \emph{International Conference on Machine Learning, {ICML} 2022, 17-23 July 2022, Baltimore, Maryland, {USA}}, volume 162 of \emph{Proceedings of Machine Learning Research}, pp.\  14139--14152. {PMLR}, 2022.
\newblock URL \url{https://proceedings.mlr.press/v162/liu22v.html}.

\bibitem[Nilsback \& Zisserman(2008)Nilsback and Zisserman]{flowers-nilsback}
Maria-Elena Nilsback and Andrew Zisserman.
\newblock Automated flower classification over a large number of classes.
\newblock \emph{Proceedings of the Indian Conference on Computer Vision, Graphics and Image Processing}, pp.\  722--729, 2008.

\bibitem[OpenAI(2023)]{GPT-4-OPENAI}
OpenAI.
\newblock {GPT-4} technical report.
\newblock \emph{CoRR}, abs/2303.08774, 2023.
\newblock \doi{10.48550/ARXIV.2303.08774}.
\newblock URL \url{https://doi.org/10.48550/arXiv.2303.08774}.

\bibitem[Parkhi et~al.(2012)Parkhi, Vedaldi, Zisserman, and Jawahar]{pets-parkhi}
Omkar~M Parkhi, Andrea Vedaldi, Andrew Zisserman, and CV~Jawahar.
\newblock Cats and dogs.
\newblock \emph{2012 IEEE Conference on Computer Vision and Pattern Recognition}, pp.\  3498--3505, 2012.

\bibitem[Paszke et~al.(2019)Paszke, Gross, Massa, Lerer, Bradbury, Chanan, Killeen, Lin, Gimelshein, Antiga, Desmaison, Kopf, Yang, DeVito, Raison, Tejani, Chilamkurthy, Steiner, Fang, Bai, and Chintala]{pytorch}
Adam Paszke, Sam Gross, Francisco Massa, Adam Lerer, James Bradbury, Gregory Chanan, Trevor Killeen, Zeming Lin, Natalia Gimelshein, Luca Antiga, Alban Desmaison, Andreas Kopf, Edward Yang, Zachary DeVito, Martin Raison, Alykhan Tejani, Sasank Chilamkurthy, Benoit Steiner, Lu~Fang, Junjie Bai, and Soumith Chintala.
\newblock Pytorch: An imperative style, high-performance deep learning library.
\newblock In \emph{Advances in Neural Information Processing Systems 32}, pp.\  8024--8035. Curran Associates, Inc., 2019.
\newblock URL \url{http://papers.neurips.cc/paper/9015-pytorch-an-imperative-style-high-performance-deep-learning-library.pdf}.

\bibitem[Razdaibiedina et~al.(2023)Razdaibiedina, Mao, Khabsa, Lewis, Hou, Ba, and Almahairi]{Residual-Prompt-Tuning-Razdaibiedina}
Anastasia Razdaibiedina, Yuning Mao, Madian Khabsa, Mike Lewis, Rui Hou, Jimmy Ba, and Amjad Almahairi.
\newblock Residual prompt tuning: improving prompt tuning with residual reparameterization.
\newblock In Anna Rogers, Jordan~L. Boyd{-}Graber, and Naoaki Okazaki (eds.), \emph{Findings of the Association for Computational Linguistics: {ACL} 2023, Toronto, Canada, July 9-14, 2023}, pp.\  6740--6757. Association for Computational Linguistics, 2023.
\newblock \doi{10.18653/V1/2023.FINDINGS-ACL.421}.
\newblock URL \url{https://doi.org/10.18653/v1/2023.findings-acl.421}.

\bibitem[Rumelhart et~al.(1986)Rumelhart, Hinton, and Williams]{rumelhart1-backprop}
David~E Rumelhart, Geoffrey~E Hinton, and Ronald~J Williams.
\newblock Learning representations by back-propagating errors.
\newblock \emph{nature}, 323\penalty0 (6088):\penalty0 533--536, 1986.

\bibitem[Singhal et~al.(2023)Singhal, Tu, Gottweis, Sayres, Wulczyn, Hou, Clark, Pfohl, Cole{-}Lewis, Neal, Schaekermann, Wang, Amin, Lachgar, Mansfield, Prakash, Green, Dominowska, y~Arcas, Tomasev, Liu, Wong, Semturs, Mahdavi, Barral, Webster, Corrado, Matias, Azizi, Karthikesalingam, and Natarajan]{singhal-parm2}
Karan Singhal, Tao Tu, Juraj Gottweis, Rory Sayres, Ellery Wulczyn, Le~Hou, Kevin Clark, Stephen Pfohl, Heather Cole{-}Lewis, Darlene Neal, Mike Schaekermann, Amy Wang, Mohamed Amin, Sami Lachgar, Philip~Andrew Mansfield, Sushant Prakash, Bradley Green, Ewa Dominowska, Blaise~Ag{\"{u}}era y~Arcas, Nenad Tomasev, Yun Liu, Renee Wong, Christopher Semturs, S.~Sara Mahdavi, Joelle~K. Barral, Dale~R. Webster, Gregory~S. Corrado, Yossi Matias, Shekoofeh Azizi, Alan Karthikesalingam, and Vivek Natarajan.
\newblock Towards expert-level medical question answering with large language models.
\newblock \emph{CoRR}, abs/2305.09617, 2023.
\newblock \doi{10.48550/ARXIV.2305.09617}.
\newblock URL \url{https://doi.org/10.48550/arXiv.2305.09617}.

\bibitem[Tan \& Le(2021)Tan and Le]{efficientnetv2-tan}
Mingxing Tan and Quoc~V. Le.
\newblock Efficientnetv2: Smaller models and faster training.
\newblock In Marina Meila and Tong Zhang (eds.), \emph{Proceedings of the 38th International Conference on Machine Learning, {ICML} 2021, 18-24 July 2021, Virtual Event}, volume 139 of \emph{Proceedings of Machine Learning Research}, pp.\  10096--10106. {PMLR}, 2021.
\newblock URL \url{http://proceedings.mlr.press/v139/tan21a.html}.

\bibitem[Taori et~al.(2023)Taori, Gulrajani, Zhang, Dubois, Li, Guestrin, Liang, and Hashimoto]{taori-insturction2-alpaca}
Rohan Taori, Ishaan Gulrajani, Tianyi Zhang, Yann Dubois, Xuechen Li, Carlos Guestrin, Percy Liang, and Tatsunori~B. Hashimoto.
\newblock Stanford alpaca: An instruction-following llama model.
\newblock \url{https://github.com/tatsu-lab/stanford_alpaca}, 2023.

\bibitem[Touvron et~al.(2023)Touvron, Martin, Stone, Albert, Almahairi, Babaei, Bashlykov, Batra, Bhargava, Bhosale, Bikel, Blecher, Canton{-}Ferrer, Chen, Cucurull, Esiobu, Fernandes, Fu, Fu, Fuller, Gao, Goswami, Goyal, Hartshorn, Hosseini, Hou, Inan, Kardas, Kerkez, Khabsa, Kloumann, Korenev, Koura, Lachaux, Lavril, Lee, Liskovich, Lu, Mao, Martinet, Mihaylov, Mishra, Molybog, Nie, Poulton, Reizenstein, Rungta, Saladi, Schelten, Silva, Smith, Subramanian, Tan, Tang, Taylor, Williams, Kuan, Xu, Yan, Zarov, Zhang, Fan, Kambadur, Narang, Rodriguez, Stojnic, Edunov, and Scialom]{llama2-tourvron}
Hugo Touvron, Louis Martin, Kevin Stone, Peter Albert, Amjad Almahairi, Yasmine Babaei, Nikolay Bashlykov, Soumya Batra, Prajjwal Bhargava, Shruti Bhosale, Dan Bikel, Lukas Blecher, Cristian Canton{-}Ferrer, Moya Chen, Guillem Cucurull, David Esiobu, Jude Fernandes, Jeremy Fu, Wenyin Fu, Brian Fuller, Cynthia Gao, Vedanuj Goswami, Naman Goyal, Anthony Hartshorn, Saghar Hosseini, Rui Hou, Hakan Inan, Marcin Kardas, Viktor Kerkez, Madian Khabsa, Isabel Kloumann, Artem Korenev, Punit~Singh Koura, Marie{-}Anne Lachaux, Thibaut Lavril, Jenya Lee, Diana Liskovich, Yinghai Lu, Yuning Mao, Xavier Martinet, Todor Mihaylov, Pushkar Mishra, Igor Molybog, Yixin Nie, Andrew Poulton, Jeremy Reizenstein, Rashi Rungta, Kalyan Saladi, Alan Schelten, Ruan Silva, Eric~Michael Smith, Ranjan Subramanian, Xiaoqing~Ellen Tan, Binh Tang, Ross Taylor, Adina Williams, Jian~Xiang Kuan, Puxin Xu, Zheng Yan, Iliyan Zarov, Yuchen Zhang, Angela Fan, Melanie Kambadur, Sharan Narang, Aur{\'{e}}lien Rodriguez, Robert Stojnic, Sergey Edunov,
  and Thomas Scialom.
\newblock Llama 2: Open foundation and fine-tuned chat models.
\newblock \emph{CoRR}, abs/2307.09288, 2023.
\newblock \doi{10.48550/ARXIV.2307.09288}.
\newblock URL \url{https://doi.org/10.48550/arXiv.2307.09288}.

\bibitem[Vaswani et~al.(2017)Vaswani, Shazeer, Parmar, Uszkoreit, Jones, Gomez, Kaiser, and Polosukhin]{Vaswani-transformer}
Ashish Vaswani, Noam Shazeer, Niki Parmar, Jakob Uszkoreit, Llion Jones, Aidan~N. Gomez, Lukasz Kaiser, and Illia Polosukhin.
\newblock Attention is all you need.
\newblock In Isabelle Guyon, Ulrike von Luxburg, Samy Bengio, Hanna~M. Wallach, Rob Fergus, S.~V.~N. Vishwanathan, and Roman Garnett (eds.), \emph{Advances in Neural Information Processing Systems 30: Annual Conference on Neural Information Processing Systems 2017, December 4-9, 2017, Long Beach, CA, {USA}}, pp.\  5998--6008, 2017.
\newblock URL \url{https://proceedings.neurips.cc/paper/2017/hash/3f5ee243547dee91fbd053c1c4a845aa-Abstract.html}.

\bibitem[Wang et~al.(2023)Wang, Wu, Dabral, Zhang, Brown, Lu, Liu, Liang, Pang, Bendersky, et~al.]{Input-Prompt-Wang}
Yaqing Wang, Jialin Wu, Tanmaya Dabral, Jiageng Zhang, Geoff Brown, Chun-Ta Lu, Frederick Liu, Yi~Liang, Bo~Pang, Michael Bendersky, et~al.
\newblock Non-intrusive adaptation: Input-centric parameter-efficient fine-tuning for versatile multimodal modeling.
\newblock \emph{arXiv preprint arXiv:2310.12100}, 2023.

\bibitem[Wang et~al.(2016)Wang, Yang, Melhem, Childers, Zhang, and Guo]{wang-multikernel1}
Zhenning Wang, Jun Yang, Rami Melhem, Bruce Childers, Youtao Zhang, and Minyi Guo.
\newblock Simultaneous multikernel gpu: Multi-tasking throughput processors via fine-grained sharing.
\newblock In \emph{2016 IEEE international symposium on high performance computer architecture (HPCA)}, pp.\  358--369. IEEE, 2016.

\bibitem[Wei et~al.(2022)Wei, Bosma, Zhao, Guu, Yu, Lester, Du, Dai, and Le]{Wei-instruction1-flan}
Jason Wei, Maarten Bosma, Vincent~Y. Zhao, Kelvin Guu, Adams~Wei Yu, Brian Lester, Nan Du, Andrew~M. Dai, and Quoc~V. Le.
\newblock Finetuned language models are zero-shot learners.
\newblock In \emph{ICLR, virtual, April 25-29, 2022}. OpenReview.net, 2022.

\bibitem[Wolf et~al.(2019)Wolf, Debut, Sanh, Chaumond, Delangue, Moi, Cistac, Rault, Louf, Funtowicz, and Brew]{huggingface}
Thomas Wolf, Lysandre Debut, Victor Sanh, Julien Chaumond, Clement Delangue, Anthony Moi, Pierric Cistac, Tim Rault, R{\'{e}}mi Louf, Morgan Funtowicz, and Jamie Brew.
\newblock Huggingface's transformers: State-of-the-art natural language processing.
\newblock \emph{CoRR}, abs/1910.03771, 2019.
\newblock URL \url{http://arxiv.org/abs/1910.03771}.

\bibitem[Woo \& Jeon(2023)Woo and Jeon]{AAL-Woo}
Sunghyeon Woo and Dongsuk Jeon.
\newblock Learning with auxiliary activation for memory-efficient training.
\newblock In \emph{The Eleventh International Conference on Learning Representations, {ICLR} 2023, Kigali, Rwanda, May 1-5, 2023}. OpenReview.net, 2023.
\newblock URL \url{https://openreview.net/forum?id=YgC62m4CY3r}.

\bibitem[Woo et~al.(2024)Woo, Lee, and Jeon]{ALAM-Woo}
Sunghyeon Woo, Sunwoo Lee, and Dongsuk Jeon.
\newblock {ALAM:} averaged low-precision activation for memory-efficient training of transformer models.
\newblock In \emph{The Twelfth International Conference on Learning Representations, {ICLR} 2024, Vienna, Austria, May 7-11, 2024}. OpenReview.net, 2024.
\newblock URL \url{https://openreview.net/forum?id=OfXqQ5TRwp}.

\bibitem[Wu et~al.(2024)Wu, Wang, Zhao, and Wong]{MosLoRA-Wu}
Taiqiang Wu, Jiahao Wang, Zhe Zhao, and Ngai Wong.
\newblock Mixture-of-subspaces in low-rank adaptation.
\newblock \emph{CoRR}, abs/2406.11909, 2024.
\newblock \doi{10.48550/ARXIV.2406.11909}.
\newblock URL \url{https://doi.org/10.48550/arXiv.2406.11909}.

\bibitem[Xu et~al.(2024)Xu, Xie, Gu, Chen, Chang, Zhang, Chen, Zhang, and Tian]{Xu-QALoRA}
Yuhui Xu, Lingxi Xie, Xiaotao Gu, Xin Chen, Heng Chang, Hengheng Zhang, Zhengsu Chen, Xiaopeng Zhang, and Qi~Tian.
\newblock Qa-lora: Quantization-aware low-rank adaptation of large language models.
\newblock In \emph{The Twelfth International Conference on Learning Representations, {ICLR} 2024, Vienna, Austria, May 7-11, 2024}. OpenReview.net, 2024.
\newblock URL \url{https://openreview.net/forum?id=WvFoJccpo8}.

\bibitem[Zhang et~al.(2024)Zhang, Han, Liu, Zhou, Lu, Qiao, Li, and Gao]{LLaMA-Adapter-Zhang}
Renrui Zhang, Jiaming Han, Chris Liu, Aojun Zhou, Pan Lu, Yu~Qiao, Hongsheng Li, and Peng Gao.
\newblock Llama-adapter: Efficient fine-tuning of large language models with zero-initialized attention.
\newblock In \emph{The Twelfth International Conference on Learning Representations, {ICLR} 2024, Vienna, Austria, May 7-11, 2024}. OpenReview.net, 2024.
\newblock URL \url{https://openreview.net/forum?id=d4UiXAHN2W}.

\bibitem[Zheng et~al.(2023)Zheng, Chiang, Sheng, Zhuang, Wu, Zhuang, Lin, Li, Li, Xing, Zhang, Gonzalez, and Stoica]{mt-bench-zheng}
Lianmin Zheng, Wei{-}Lin Chiang, Ying Sheng, Siyuan Zhuang, Zhanghao Wu, Yonghao Zhuang, Zi~Lin, Zhuohan Li, Dacheng Li, Eric~P. Xing, Hao Zhang, Joseph~E. Gonzalez, and Ion Stoica.
\newblock Judging llm-as-a-judge with mt-bench and chatbot arena.
\newblock In Alice Oh, Tristan Naumann, Amir Globerson, Kate Saenko, Moritz Hardt, and Sergey Levine (eds.), \emph{Advances in Neural Information Processing Systems 36: Annual Conference on Neural Information Processing Systems 2023, NeurIPS 2023, New Orleans, LA, USA, December 10 - 16, 2023}, 2023.
\newblock URL \url{http://papers.nips.cc/paper\_files/paper/2023/hash/91f18a1287b398d378ef22505bf41832-Abstract-Datasets\_and\_Benchmarks.html}.

\end{thebibliography}
\bibliographystyle{iclr2025_conference}

\newpage
\appendix
\section*{Appendices}
\section{Proof for Convergence of PaCA}\label{appendix: proof for the convergence of PaCA}

\setcounter{theorem}{0}
\begin{theorem} 
If the gradient of the loss function $f(\textbf{W}, \textbf{X})$ is Lipschitz continuous and the only partial connections are updated, then 
\begin{equation*}
    f(\textbf{W}^{k+1}, \textbf{X}^{k+1}) \leq f(\textbf{W}^{k}, \textbf{X}^{k}) -\eta(1-\frac{\eta L}{2}) ||\nabla \textbf{P}^{k}||^{2}
\end{equation*}
\end{theorem}

\begin{proof}
    As the gradient of the loss function $f(\textbf{W}, \textbf{X})$ is Lipschitz continuous, we obtain

    \begin{align*}
        f(\textbf{W}^{k+1}, \textbf{X}^{k+1}) \leq f(\textbf{W}^{k}, \textbf{X}^{k}) + \nabla_{\textbf{W}^{k}}f(\textbf{W}^{k}, \textbf{X}^{k})^{T}(\textbf{W}^{k+1}-\textbf{W}^{k})+\frac{L}{2}||\textbf{W}^{k+1}-\textbf{W}^{k}||^{2}
    \end{align*}

    By substituting Eq. \ref{eq: paca weight update} which represents partial connection updates, we obtain

    \begin{align*}
        f(\textbf{W}^{k+1}, \textbf{X}^{k+1}) &\leq f(\textbf{W}^{k}, \textbf{X}^{k}) + \nabla_{\textbf{W}^{k}}f(\textbf{W}^{k}, \textbf{X}^{k})(\textbf{W}^{k+1}-\textbf{W}^{k})^{T}+\frac{L}{2}||\textbf{W}^{k+1}-\textbf{W}^{k}||^{2}\\
         &= f(\textbf{W}^{k}, \textbf{X}^{k}) + \nabla_{\textbf{W}^{k}}f(\textbf{W}^{k}, \textbf{X}^{k})(-\eta \Delta \textbf{W}^{k})^{T}+\frac{L}{2}||-\eta \Delta \textbf{W}^{k}||^{2}\\
         &= f(\textbf{W}^{k}, \textbf{X}^{k}) - \eta (\nabla_{\textbf{W}^{k}}f(\textbf{W}^{k}, \textbf{X}^{k}) - \frac{\eta L}{2} \Delta \textbf{W}^{k})(\Delta \textbf{W}^{k})^{T} \\
         &= f(\textbf{W}^{k}, \textbf{X}^{k}) - \sum_{l=1}^{n}\eta(\nabla_{\textbf{W}_{l}^{k}}f(\textbf{W}^{k}, \textbf{X}^{k}) - \frac{\eta L}{2} \Delta \textbf{W}_{l}^{k})(\Delta \textbf{W}_{l}^{k})^{T} \\
         &= f(\textbf{W}^{k}, \textbf{X}^{k}) - \sum_{l=1}^{n}\eta(\nabla_{\textbf{W}_{l}^{k}}f(\textbf{W}^{k}, \textbf{X}^{k}) - \frac{\eta L}{2} \Delta \textbf{W}_{l}^{k})(\Delta \textbf{W}_{l}^{k})^{T}
    \end{align*}

    Also, $\nabla_{\textbf{W}_{l}^{k}}f(\textbf{W}^{k}, \textbf{X}^{k})$ and $\Delta \textbf{W}_{l}^{k}$ can be expressed as

    \begin{align*}
        \nabla_{\textbf{W}_{l}^{k}}f(\textbf{W}^{k}, \textbf{X}^{k}) &= \left[ {}_m\nabla w_l^k \right]_{m=1}^{d_{l}} \\
        \Delta \textbf{W}_{l}^{k} &=  \left[ {}_m\nabla w_l^k \ \text{if} \ m \in I=\{i_1, i_2, \dots, i_r\}, \ \text{else} \ \textbf{0} \right]_{m=1}^{d_{l}}
    \end{align*}

    where $I$ represents the set of indices corresponding to the selected columns. By applying $\nabla_{\textbf{W}_{l}^{k}}f(\textbf{W}^{k}, \textbf{X}^{k})$ and $\Delta \textbf{W}_{l}^{k}$ above, we obtain

      \begin{align*}
        f(\textbf{W}^{k+1}, & \textbf{X}^{k+1}) \leq f(\textbf{W}^{k}, \textbf{X}^{k}) - \sum_{l=1}^{n}\eta(\nabla_{\textbf{W}_{l}^{k}}f(\textbf{W}^{k}, \textbf{X}^{k}) - \frac{\eta L}{2} \Delta \textbf{W}_{l}^{k})(\Delta \textbf{W}_{l}^{k})^{T} \\
        &=  f(\textbf{W}^{k}, \textbf{X}^{k}) - \sum_{l=1}^{n}\eta \left[ (1-\frac{\eta L}{2}) {}_m\nabla w_l^k \ \text{if} \ m \in I, \ \text{else} \ {}_m\nabla w_l^k \right]_{m=1}^{d_{l}} (\Delta \textbf{W}_{l}^{k})^T\\
        &=  f(\textbf{W}^{k}, \textbf{X}^{k}) -  \sum_{l=1}^{n} \sum_{m \in I}\eta(1-\frac{\eta L}{2}) ||{}_m\nabla w_l^k ||^2\\
        &= f(\textbf{W}^{k}, \textbf{X}^{k}) -  \sum_{l=1}^{n} \eta(1-\frac{\eta L}{2}) ||\nabla\textbf{P}_l^k||^2
        = f(\textbf{W}^{k}, \textbf{X}^{k}) -  \eta(1-\frac{\eta L}{2}) ||\nabla\textbf{P}^k||^2
    \end{align*}

\end{proof}

\textcolor{black}{We assumed the Lipschitz continuity of gradients to theoretically prove the convergence of PaCA. However, we acknowledge the inherent limitations of the Lipschitz continuity assumption. In practice, this assumption may not hold for certain neural networks, particularly in scenarios where gradient magnitudes vary significantly due to sharp activation functions, high model complexity, or specific architectural designs. It is well known that it is very challenging to theoretically analyze the convergence of general deep neural networks. Therefore, prior studies \citep{DGL-Beliovsky, ActNN-Chen, GACT-Liu, AAL-Woo} first proved the convergence of the proposed algorithm under weak constraints, such as the Lipschitz continuity of gradients, and then validated convergence empirically in real-world scenarios.}

\textcolor{black}{Following a similar approach, we assumed the Lipschitz continuity of gradients to theoretically prove the convergence of PaCA. Then, we experimentally demonstrated that PaCA successfully trains real-world large-scale neural networks such as LLaMA Models, where the Lipschitz continuity assumption may not strictly hold, as shown in Tables \ref{tb: task specific fine-tuning}-\ref{tb: qpaca} in Section \ref{sec: experiments}. }

\section{\textcolor{black}{Applicability of PaCA to Other Architectures and Tasks}}\label{appendix: vision}
\textcolor{black}{In this section, we fine-tune ViT-B/16 \citep{vit-dosovitskiy} and EfficientNetV2-L \citep{efficientnetv2-tan} using various datasets such as CIFAR-10 \citep{cifar10-krizhevsky}, CIFAR-100 \citep{cifar10-krizhevsky}, Oxford-IIIT Pets \citep{pets-parkhi}, and Oxford-Flowers 102 \citep{flowers-nilsback} to evaluate the generalizability of PaCA.} 

\begin{table}[!ht]
    \centering
    \caption{\textcolor{black}{Comparisons of memory usage (Mem), training time (Time), and accuracy when fine-tuning ViT-B/16 on CIFAR-10, CIFAR-100, Oxford-III Pets, and Oxford-Flowers 102.}}
    \label{tb:vit_table}
    \begin{tabular}{@{}c|c|c|cccc|c@{}}
        \toprule
        \multicolumn{1}{c|}{\multirow{2}{*}{Method}} & 
        \multicolumn{1}{c|}{\multirow{2}{*}{Mem}} & 
        \multicolumn{1}{c|}{\multirow{2}{*}{Time}} & 
        \multicolumn{5}{c}{Accuracy (\%)}  \\ 
        \cmidrule(lr){4-8}
        & & & CIFAR10 & CIFAR100 & IIIT Pets & Flowers102 & Avg. \\ 
        \midrule
        LoRA & 11.0G & 45m & 98.9 & 92.5 & 93.6 & 99.2 & 96.1 \\
        \rowcolor{blue!10} PaCA (Ours) & \textbf{6.7G} & \textbf{32m} & 98.9 & 92.8 & 93.9 & 99.1 & \textbf{96.2} \\
        \bottomrule
    \end{tabular}
\end{table}

\begin{table}[!ht]
    \centering
    \caption{\textcolor{black}{Comparisons of memory usage (Mem), training time (Time), and accuracy when fine-tuning EfficientNetV2-L on CIFAR-10 and CIFAR-100.}}
    \label{tb:efficientnetv2}
    \begin{tabular}{@{}c|c|c|cc|c@{}}
        \toprule
        \multicolumn{1}{c|}{\multirow{2}{*}{Method}} & 
        \multicolumn{1}{c|}{\multirow{2}{*}{Mem}} & 
        \multicolumn{1}{c|}{\multirow{2}{*}{Time}} & 
        \multicolumn{3}{c}{Accuracy (\%)} \\ 
        \cmidrule(lr){4-6}
        & & & CIFAR10 & CIFAR100 & Avg. \\ 
        \midrule
        Full-FT & 18.3 GB & 70m & 98.5 & 90.1 & \textbf{94.3} \\
        \rowcolor{blue!10} PaCA (Ours) & \textbf{13.2 GB} & \textbf{59m} & 98.0 & 89.3 & 93.7 \\
        \bottomrule
    \end{tabular}
\end{table}

\textcolor{black}{Table \ref{tb:vit_table} shows that our PaCA achieves comparable accuracy to LoRA while reducing training memory and time by 39\% and 29\%, respectively, on the ViT-B/16 model. Similarly, in Table \ref{tb:efficientnetv2}, PaCA demonstrated its effectiveness on EfficientNetV2-L, achieving comparable accuracy while saving 28\% in training memory and 16\% in training time compared to full fine-tuning.} 

\textcolor{black}{It should be noted that conventional PEFT algorithms such as LoRA face critical limitations when applied to convolutional neural networks since the additional adapters in LoRA are implemented as linear layers, which makes it impossible to directly merge them into a pretrained layer in a different type (e.g., convolutional layer) during inference. In contrast, PaCA fine-tunes a subset of the existing pretrained weights, enabling seamless applications to diverse types of layers including convolutional layers, ensuring its generalizability.}

\newpage

\section{Experimental Details}\label{appendix: experimental details}
\begin{table}[!hbt]
\caption{Hyperparameters used for analyzing the number of operations and the average training time per iteration, averaged over 100 iterations, for fine-tuning LLaMA3-8B.}
\centering
\label{tb: hyperparameters, FLOPs vs actual time}
\renewcommand{\arraystretch}{1.1} 
\begin{tabular}{c|ccc}
\toprule
Hyperparameters       & Full-FT           & LoRA          & PaCA         \\ \midrule
Training Precision    & \multicolumn{3}{c}{16 bits}                       \\ 
Rank                  & \multicolumn{3}{c}{8}                            \\ 
Batch   Size per Step & \multicolumn{3}{c}{2}                            \\ 
Sequence   Length     & \multicolumn{3}{c}{512}                          \\ 
Target   Modules      & \multicolumn{3}{c}{Q,   K, V, O, Up, Down, Gate} \\ \bottomrule
\end{tabular}
\end{table}

\begin{table}[!hbt]
\caption{\textcolor{black}{Hyperparameters when fine-tuning LLaMA2-7B/13B and LLaMA3-8B using PEFT algorithms on the MMLU dataset.}}
\centering
\small
\label{tb: hyperparameters, mmlu}
\renewcommand{\arraystretch}{1.1} 
\begin{tabular}{c|cccc}
\toprule
\multicolumn{1}{c|}{Hyperparameters} & LoRA      & DoRA      & MosLoRA    & PaCA        \\ \midrule
Rank                                 & 8 & 8 & 8              & 8/   16     \\
$\alpha$                             & 32 & 32 & 32             & 32/   64    \\
DropOut                              & 0.1 & 0.1 & 0.1            & -           \\
LR   (LLaMA2-7B)                 & 3e-4 & 3e-4 & 3e-4 & 3e-4/ 1e-4           \\
LR   (LLaMA2-13B)                 & 2e-4 & 1e-4 & 1e-4 & 1e-4/ 1e-4           \\
LR   (LLaMA3-8B)                     & 1e-5 & 5e-6 & 5e-6 & 5e-6/ 5e-6                   \\
Training Precision                   & \multicolumn{4}{c}{16-bit mixed precision}       \\
Optimizer                            & \multicolumn{4}{c}{AdamW}                        \\
LR   Scheduler                       & \multicolumn{4}{c}{cosine}                       \\
Batch Size                           & \multicolumn{4}{c}{8}                            \\
Gradient   Accumulation Steps        & \multicolumn{4}{c}{4}                            \\
Sequence   Length                    & \multicolumn{4}{c}{512}                          \\
Warmup   Steps                       & \multicolumn{4}{c}{100}                          \\
Epochs                               & \multicolumn{4}{c}{1}                            \\
Target   Modules                     & \multicolumn{4}{c}{Q,   K, V, O, Up, Down, Gate} \\ \bottomrule
\end{tabular}
\end{table}

\begin{table}[!hbt]
\caption{\textcolor{black}{Hyperparameters used when fine-tuning LLaMA3-8B using PEFT algorithms on the Oasst1 dataset.}}
\centering
\small
\label{tb: hyperparameters, oasst1}
\renewcommand{\arraystretch}{1.1} 
\begin{tabular}{c|cccc}
\toprule
Hyperparameters               & LoRA      & DoRA     & MosLoRA    & PaCA         \\ \midrule
Rank                          &64 & 64 & 64            & 64/   128    \\
$\alpha$                      &1 & 1 & 1 &  1                           \\
DropOut                       & \multicolumn{4}{c}{-}                         \\
Training Precision            & \multicolumn{4}{c}{16-bit mixed precision}       \\
Optimizer                     & \multicolumn{4}{c}{AdamW}                        \\
LR                 & \multicolumn{4}{c}{5e-4,   1e-3, 5e-3}           \\
LR   Scheduler                & \multicolumn{4}{c}{linear}                       \\
Batch Size                    & \multicolumn{4}{c}{16}                           \\
Gradient   Accumulation Steps & \multicolumn{4}{c}{4}                            \\
Sequence   Length             & \multicolumn{4}{c}{768}                          \\
Warmup   Ratio                & \multicolumn{4}{c}{0.1}                          \\
Epochs                        & \multicolumn{4}{c}{1}                            \\ 
Target   Modules              & \multicolumn{4}{c}{Q,   K, V, O, Up, Down, Gate} \\ \bottomrule
\end{tabular}
\end{table}

\begin{table}[H]
\caption{\textcolor{black}{Hyperparameters used  when fine-tuning LLaMA3.1-70B using QLoRA and QPaCA on the Oasst1 dataset.}}
\centering
\small
\label{tb: hyperparameters, qlora}
\renewcommand{\arraystretch}{1.1} 
\begin{tabular}{ccc}
\toprule
Hyperparameters             & LLaMA-8B                   & LLaMA3.1-70B                  \\ \midrule
Gradient Accumulation Steps & 4 & 2                            \\
Rank                        & \multicolumn{2}{c}{64}                           \\
$\alpha$                    & \multicolumn{2}{c}{1}                            \\
DropOut                     & \multicolumn{2}{c}{-}                      \\
Training Precision          & \multicolumn{2}{c}{16-bit mixed precision}       \\
Optimizer                   & \multicolumn{2}{c}{AdamW}                        \\
LR                          & \multicolumn{2}{c}{5e-4, 1e-3, 5e-3}             \\
LR   Scheduler              & \multicolumn{2}{c}{linear}                       \\
Batch Size                  & \multicolumn{2}{c}{16}                           \\
Sequence   Length           & \multicolumn{2}{c}{768}                          \\
Warmup   Ratio              & \multicolumn{2}{c}{0.1}                          \\
Epochs                      & \multicolumn{2}{c}{1}                            \\
Target   Modules            & \multicolumn{2}{c}{Q,   K, V, O, Up, Down, Gate} \\ \bottomrule
\end{tabular}
\end{table}

\begin{table}[H]
\caption{Hyperparameters used for verifying the maximum sequence length on a single GPU for fine-tuning LLaMA3-8B.}
\centering
\label{tb: hyperparameters, max length}
\renewcommand{\arraystretch}{1.1} 
\begin{tabular}{cccccc}
\toprule
Hyperparameters         & Full-FT   & LoRA   & DoRA   & MosLoRA   & PaCA   \\ \midrule
Rank                    & \multicolumn{5}{c}{8}                            \\
Training Precision      & \multicolumn{5}{c}{16-bit mixed precision}                       \\
Batch   Size   per Step & \multicolumn{5}{c}{1}                            \\
Target   Modules        & \multicolumn{5}{c}{Q,   K, V, O, Up, Down, Gate} \\ \bottomrule
\end{tabular}
\end{table}

\begin{table}[H]
\caption{Hyperparameters for comparing training throughput when increasing batch size on a single GPU for fine-tuning LLaMA3-8B.}
\centering
\label{tb: hyperparameters, training throughput}
\renewcommand{\arraystretch}{1.1} 
\begin{tabular}{cccccc}
\toprule
Hyperparameters    & Full-FT   & LoRA   & DoRA   & MosLoRA   & PaCA   \\ \midrule
Rank               & \multicolumn{5}{c}{8}                            \\
Training Precision & \multicolumn{5}{c}{16-bit}                       \\
Sequence Length    & \multicolumn{5}{c}{512}                          \\
Target   Modules   & \multicolumn{5}{c}{Q,   K, V, O, Up, Down, Gate} \\ \bottomrule
\end{tabular}
\end{table}

\end{document}